\documentclass[sigconf]{acmart}
\graphicspath{{figures/}}
\usepackage[ruled,vlined,linesnumbered]{algorithm2e}
\usepackage{graphicx}
\usepackage{subcaption}
\usepackage{enumitem}
\usepackage{multirow}
\usepackage{makecell}
\usepackage[symbol]{footmisc}
\AtBeginDocument{%
  \providecommand\BibTeX{{%
    \normalfont B\kern-0.5em{\scshape i\kern-0.25em b}\kern-0.8em\TeX}}}

\acmConference[KDD '22]{KDD '22: 28th ACM SIGKDD Conference on Knowledge Discovery and Data Mining (KDD ’22)}{August 14-18, 2022}{Washington}
\acmBooktitle{28th ACM SIGKDD Conference on Knowledge Discovery and Data Mining (KDD ’22),
  August 14-18, 2022, Washington}
\acmPrice{xx.00}
\acmISBN{978-1-4503-XXXX-X/18/06}

\begin{document}

\title{A Framework for Multi-stage Bonus Allocation in meal delivery Platform}


% \author{Zhuolin Wu, Li Wang*, Fangsheng Huang}

%  \email{{wuzhuolin,wangli78,huangfangsheng,yechengpeng,helong04,niepengyu,renhao05,haojinghua,herenqing,sunzhizhao}@meituan.com}

% \affiliation{%
%   \institution{Meituan Group}
%   \city{Beijing}
%   \country{China}
% }

\author{Zhuolin Wu}
\email{sleeplinzw@163.com}
\affiliation{%
  \institution{Meituan Group}
  \city{Beijing}
  \country{China}
}

\author{Li Wang}
\authornote{Corresponding author}
\email{liliw27@gmail.com}
\affiliation{%
  \institution{Huazhong University of Science and Technology}
  \city{Wuhan}
  \country{China}
}

\author{Fangsheng Huang, Linjun Zhou, Yu Song, Chengpeng Ye, Pengyu Nie, Hao Ren, Jinghua Hao, Renqing He, Zhizhao Sun}
\authornote{email: \{huangfangsheng,zhoulinjun,songyu20,yechengpeng,niepengyu,renhao05,
haojinghua,herenqing,sunzhizhao\}@meituan.com}
%% \email{{huangfangsheng,zhoulinjun,songyu20,yechengpeng,niepengyu,renhao05,
%% haojinghua,herenqing,sunzhizhao}@meituan.com}
\affiliation{%
  \institution{Meituan Group}
  \city{Beijing}
  \country{China}
}

 \renewcommand{\shortauthors}{Zhuolin Wu and Li Wang, et al.}

\begin{abstract}
Online meal delivery is undergoing explosive growth, as this service is becoming increasingly popular. A meal delivery platform aims to provide excellent and stable services for customers and restaurants. However, in reality, several hundred thousand orders are canceled per day in the Meituan meal delivery platform since they are not accepted by the crowd soucing drivers. The cancellation of the orders is incredibly detrimental to the customer's repurchase rate and the reputation of the Meituan meal delivery platform. To solve this problem, a certain amount of specific funds is provided by Meituan's business managers to encourage the crowdsourcing drivers to accept more orders. To make better use of the funds, in this work, we propose a framework to deal with the multi-stage bonus allocation problem for a meal delivery platform. The objective of this framework is to maximize the number of accepted orders within a limited bonus budget. This framework consists of a semi-black-box acceptance probability model, a Lagrangian dual-based dynamic programming algorithm, and an online allocation algorithm. The semi-black-box acceptance probability model is employed to forecast the relationship between the bonus allocated to order and its acceptance probability, the Lagrangian dual-based dynamic programming algorithm aims to calculate the empirical Lagrangian multiplier for each allocation stage offline based on the historical data set, and the online allocation algorithm uses the results attained in the offline part to calculate a proper delivery bonus for each order. To verify the effectiveness and efficiency of our framework, both offline experiments on a real-world data set and online A/B tests on the Meituan meal delivery platform are conducted. Our results show that using the proposed framework, the total order cancellations can be decreased by more than 25\% in reality.
 
 %Since this acceptance probability model is non-linear and non-convex, the multistage bonus allocation problem is formulated as a non-convex optimization model. Moreover, there are three challenges in making the delivery bonus allocation decision by solving the model in real time: 1. The non-convex optimization model is intractable in practice; 2. The long-term budget constraints are challenging to be dealt with, because bonus allocation decisions need to be made in real time and total available bonus is restricted within a predetermined weekly/monthly budget. without the whole information of the orders placed during the week/month; 3. Multiple stages of bonus allocation are performed for 30 million orders everyday and each decision must be made in milliseconds. Therefore, to solve this problem effectively and efficiently, we propose an offline-online algorithm based on dynamic programming and Lagrangian dual theory. First, in the offline part, we calculate the empirical Lagrangian multiplier for each allocation stage based on the historical data set. Second, the results attained in the offline part are used to calculate a proper delivery bonus for each order in the online part. Both offline experiments on the real-world data set and online A/B tests on the Meituan meal delivery platform are conducted to verify the effectiveness and efficiency of our framework.
\end{abstract}

%   <concept>
%       <concept_id>10010405.10010481.10010484.10011817</concept_id>
%       <concept_desc>Applied computing~Multi-criterion optimization and decision-making</concept_desc>
%       <concept_significance>500</concept_significance>
%       </concept>
%   <concept>
%       <concept_id>10010405.10010481.10010484</concept_id>
%       <concept_desc>Applied computing~Decision analysis</concept_desc>
%       <concept_significance>500</concept_significance>
%       </concept>
%  </ccs2012>

% \ccsdesc[500]{Applied computing~Multi-criterion optimization and decision-making}
% \ccsdesc[500]{Applied computing~Decision analysis}

\keywords{multi-stage bonus allocation, meal delivery platform, Lagrangian dual-based dynamic programming, real-time optimization}

% \centering
%   \includegraphics[width=\columnwidth]{teaser.png}
%   \caption{The Framework of multistage Dynamic Pricing}

\maketitle

% \footnotetext{*Both authors contributed equally to this research.}

\section{Introduction}
With the explosive growth of e-commerce, online meal delivery is becoming an essential service in our daily life. For example, Meituan, the most popular Chinese meal delivery platform, takes 30 million meal orders each day. The platform aims to provide excellent and stable services for restaurants and customers.
 
\begin{figure}[ht]
  \centering
  \includegraphics[width=0.35\textwidth]{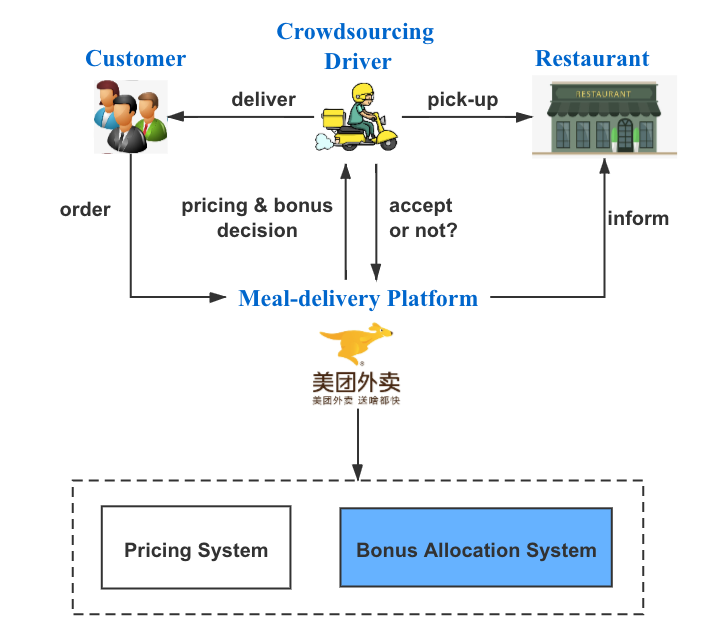}
  \caption{The typical process in the meal delivery platform.}
  \Description{need to add description}
  \label{fig:flowchart}
\end{figure}

After the customer orders a meal through the Meituan application, the corresponding order information is immediately sent to the meal delivery platform, and a typical process for the meal delivery platform is depicted in Figure \ref{fig:flowchart}. Firstly, the pricing system determines the delivery price of the order based on its properties, such as restaurant and customer locations, customer-side service difficulty, etc. Secondly, the order information, including the category, price, and estimated delivery time is pushed to nearby crowdsourcing drivers. Thirdly, a confident crowdsourcing driver accepts the order, picks up the meal from the restaurant, and delivers it to the customer. However, if the delivery price is not attractive enough, making it unaccepted for a long period, the customer might cancel the order. For simplicity, in the remaining sections, we call this kind of cancelled orders as \textit{NA-canceled orders}, \textit{i.e.} No-Accept orders.

NA-canceled order is the main reason for negative ratings of the platform, taking Meituan\footnote{\url{https://waimai.meituan.com/}} as an example, about 30,000 negative reviews are received every day, and NA-canceled orders cause over 55\% of them. Moreover, about 165,000 NA-canceled orders take place every day, which means less income for the crowdsourcing drivers, more food waste for restaurants, and a lower reputation of the meal delivery platform. The meal delivery platform needs to compensate the cost price of the restaurant's food waste due to the platform's responsibility, which is about billions of RMB for NA-canceled orders per year.
 
 %As shown in Figure \ref{fig:profit}, the order cancellation means less income for the crowdsourcing drivers, lower quality of user experience, more food waste for restaurants, and worse reputation of the meal delivery platform. Besides, the compensation is paid to the restaurant by the platform for food waste. Therefore, there is an urgent need for the bonus allocation mechanism to stimulate the crowdsourcing drivers to accept more orders.   

%  \begin{figure}[h]
%   \centering
%   \includegraphics[width=0.3\textwidth]{profit.png}
%   \caption{The impact of delivery bonus on each stakeholder}
%   \Description{need to add description}
%   \label{fig:profit}
% \end{figure}

Through analyzing the historical data, the NA-canceled orders are led by following two main reasons: On one hand, the delivery price of some orders is not attractive enough to drivers, although the number of drivers is enough to satisfy the delivery demands. On the other hand, in some cases, the number of drivers  can not enough to serve incoming orders, such as the lack of drivers online in the stormy weather. In this paper, we only focus on encouraging the driver to accept more orders in the first case. The second case, caused by extreme weather, is beyond the scope of this article.

Bonus allocation is a straightforward and effective way to encourage the drivers to accept the orders. In reality, an order is pushed to multiple nearby drivers simultaneously, and the acceptance probability of an order drops significantly with the time passing due to the potential risk of delay increasing. To motivate drivers to accept more orders and improve the efficiency of subsidy, the platform tends to allocate the bonus based on the empirical rule. For example, it allocates three RMB for the orders unaccepted within ten minutes, six RMB for twenty minutes, and so on. This method is easy to implement, it fails to achieve satisfactory performance since it offers a same price for all orders on every single decision stage separately and lacks global planning. The order's life cycle could be split to multiple decision stages, and the order might be accepted, unaccepted, or canceled at each stage. If an order is not accepted by drivers or canceled by the consumer at a particular stage, it would be transited to the next stage. Once it is accepted by a driver or canceled by the consumer at a particular stage, its life cycle stops. If no drivers take the order for more than 50 minutes, the meal delivery platform will cancel it forcibly. Although most orders would be accepted in the first few stages, there still exist a considerable number of orders unaccepted until the last few stages. Intuitively, a better bonus allocation strategy should be globally optimized based on multi-stage information.

This paper develops a Multi-Stage Bonus Allocation (MSBA) framework for the bonus allocation problem on the meal delivery platform. Zhao et al.\cite{zhao2019unified} studied a marketing budget allocation problem similar to ours. However, they solved a single-stage problem while we dealt with a multi-stage one. To the best of our knowledge, this is the first study to discuss MSBA for the meal delivery platform, which is not trivial because of the following challenges:
\begin{itemize}[leftmargin=*]
\item The bonus decided on each stage is just a display bonus, and will be paid to the driver only when the order is accepted. So this problem involves both multi-stage decision-making and complicated budget allocation.
\item In reality, the bonus allocation decision for an order must be performed in milliseconds. An efficient algorithm is proposed to generate fast and efficient bonus allocation decisions to meet such a stringent computational target.
\item The total amount of available bonuses is limited to a predetermined monthly budget, but all the orders placed within a month cannot be obtained beforehand. Therefore, solving this problem faces two challenges: 1. How to effectively use historical information to make current decisions. 2. How to dynamically adjust the strategy based on the randomly placed orders that have been generated in the current month.
\end{itemize}

The main contribution of this study is the development of an MSBA framework. This framework includes an acceptance and cancellation model, a Lagrangian dual-based dynamic programming (LDDP) algorithm, an online allocation algorithm, and a periodic control strategy. The acceptance model is employed to forecast the relationship between the bonus allocated to order and its acceptance probability. The cancellation probability is estimated for each order by an XGBoost~\cite{chen2015xgboost} model at each stage. The LDDP algorithm, which is the backbone of the framework, is developed to calculate an effective parameter (Lagrangian multiplier~\cite{yang2001nonlinear}) for each allocation stage offline based on the historical dataset. The resulting Lagrangian multipliers are used to infer the bonus allocation decisions online. Finally, we use periodic control strategies to adapt the remained budget and order set dynamically so that the cost meets the total budget constraint. Efficiently, the proposed framework can make quick decisions in real-time with the computational complexity of O(1) for online calculation. Our online A/B tests show that compared with single-stage allocation method, the number of cancelled orders reduces more than 25\%. Furthermore, we can save more than 30\% of the compensation paid to restaurants for food waste. In addition, the proposed method could be also applied to similar time-series pricing problems, such as discounts on products near expiration date in supermarkets, price design of perishable products, etc. The proposed algorithm has been already actually applied in the largest meal delivery platform in China.
\section{Problem Formulation}\label{sec:desc}

\subsection{The Acceptance and Cancellation Model}\label{sec:pro-model}
\begin{figure}[h]
  \centering
  \includegraphics[width=0.48\textwidth]{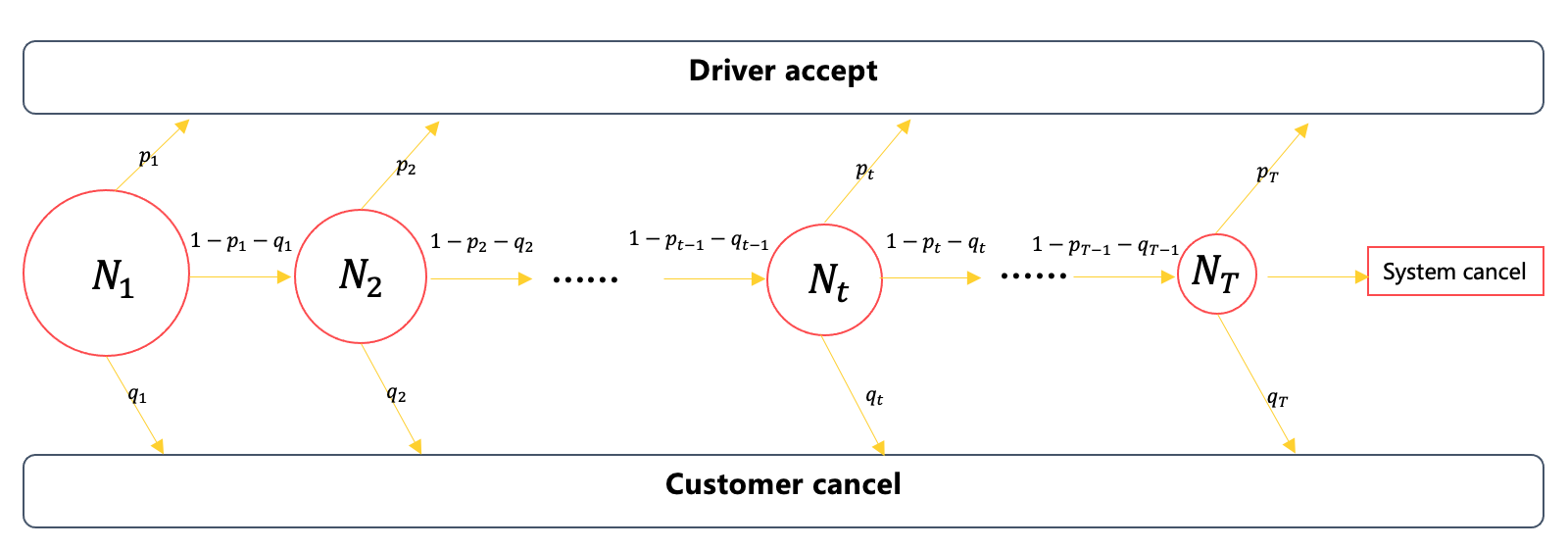}
  \caption{The transition process of an order.}
  \Description{need to add description}
  \label{fig:ordertran}
\end{figure}

 As previously mentioned, MSBA aims to maximize the number of accepted orders. To formulate this, our objective function is to maximize the expected value of the number of accepted orders based on the acceptance probability of orders. With multiple allocation stages, an unaccepted and not canceled order from a former allocation stage is transited to the next allocation stage. The transition process of the order is illustrated in Figure \ref{fig:ordertran}. Each node represents an allocation stage. Let $p_1$ be the acceptance probability in the first allocation stage,$q_1$ be the cancellation probability in the first allocation stage. Correspondingly, $1-p_1-q_1$ is the probability of transitioning to the second allocation stage. Consequently, the probability that an order can enter   $|T|$  stage is $\prod_{t_0=1}^{|T-1|}(1-p_{t_0}-q_{t_0})$. Otherwise, the order is 
accepted in  $|T|$  stage with the probability of $\prod_{t_0=1}^{|T-1|}(1-p_{t_0}-q_{t_0})p_{|T|}$. In total, the probability of an order being accepted is expressed as follows:
\begin{equation}
    p=\sum_{t=1}^{|T|}\prod_{t_0=1}^{t-1}(1-p_{t_0}-q_{t_0})p_{|T|}
\end{equation}

We need to forecast the relationship between the bonus allocated to orders and its acceptance probability to make optimal delivery bonus allocation decisions. To simplify the calculation, we assume that if the order is still on the waiting list, the probability of acceptance of the order at stage $t$ is only determined by the price of the order at stage $t$. The  price of the order at other stages and the price of other orders do not affect the acceptance probability of this order.

\begin{figure}[h]
  \centering
  \includegraphics[width=0.30\textwidth]{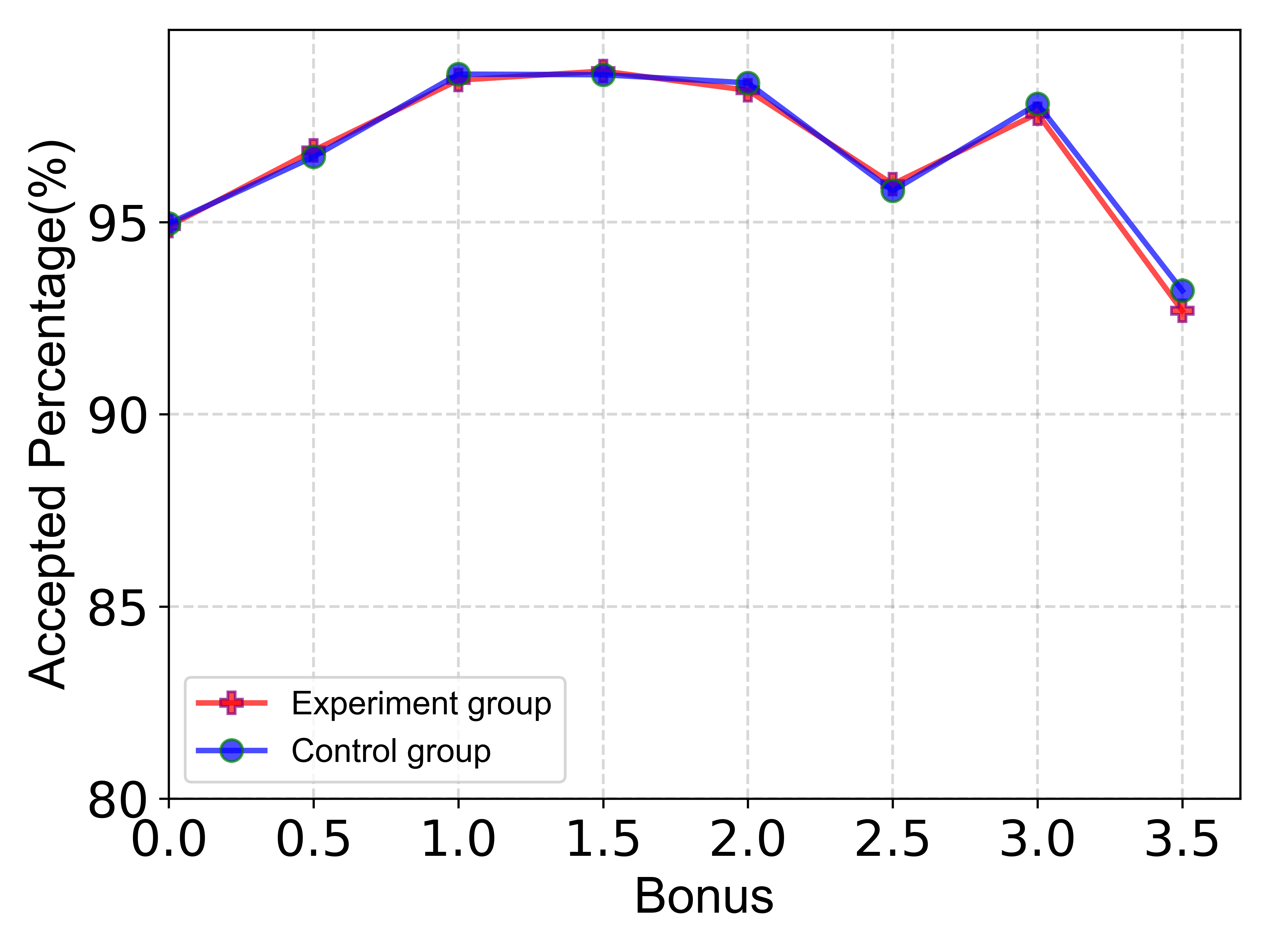}
  \caption{The proportion of orders accepted within 10 minutes.}
  \Description{need to add description}
  \label{fig:ordertran2}
\end{figure}

To test this hypothesis in real cases, we further conduct an ablation study. Orders are divided into the experiment group and the control group. We give a fixed bonus changing from 0.0 to 3.5 if the orders are not picked up within 10 minutes in the experiment group, while there is no bonus in the control group. Figure \ref{fig:ordertran2} shows the accepted rate of the orders \textbf{within 10 minutes} for different bonus level. It shows that the strategy of bonus after 10 minutes neither affects the drivers' decisions within 10 minutes, nor causes the drivers to wait for bonus, thus the accepted rate of the two groups within 10 minutes is nearly the same. Hence, the hypothesis is to some extent valid in our real delivery system.

%Although black-box forecasting methods, such as neural networks, are widely used in many applications, there are still gaps between the black-box model and bonus allocation decision-making. Instead, 
In this study, we adopt a semi-black-box forecasting model. We assume that the acceptance probability model conforms to the following logistic function:
\begin{align}
    \label{eq1.4} & p_{i,t}(c_{i,t})=\frac{1}{1+e^{\alpha_{i,t} c_{i,t}+\beta_{i,t}}}
\end{align}

Meanwhile, $\alpha_{i,t}$ and $\beta_{i,t}$ are attained by the machine learning model such as a neural network. The input features of the forecasting model can be split into two different parts: the bonus allocated to each order $c_{i,t}$ and contextual features $\boldsymbol{x_{i,t}}$. Contextual features are intrinsic attributes of the order, including the geographical locations of the customer and the restaurant, the time difference between the current time and the user's order, estimated time of arrival (ETA), the influence of the supply and demand related to drivers, the drivers’ spatial information (\textit{e.g.} the number of drivers within 2 kilometers of the restaurant), and others. The contextual features contain information as much as available unless prohibited by local law.

\begin{figure}[ht]
  \centering
  \includegraphics[width=0.3\textwidth]{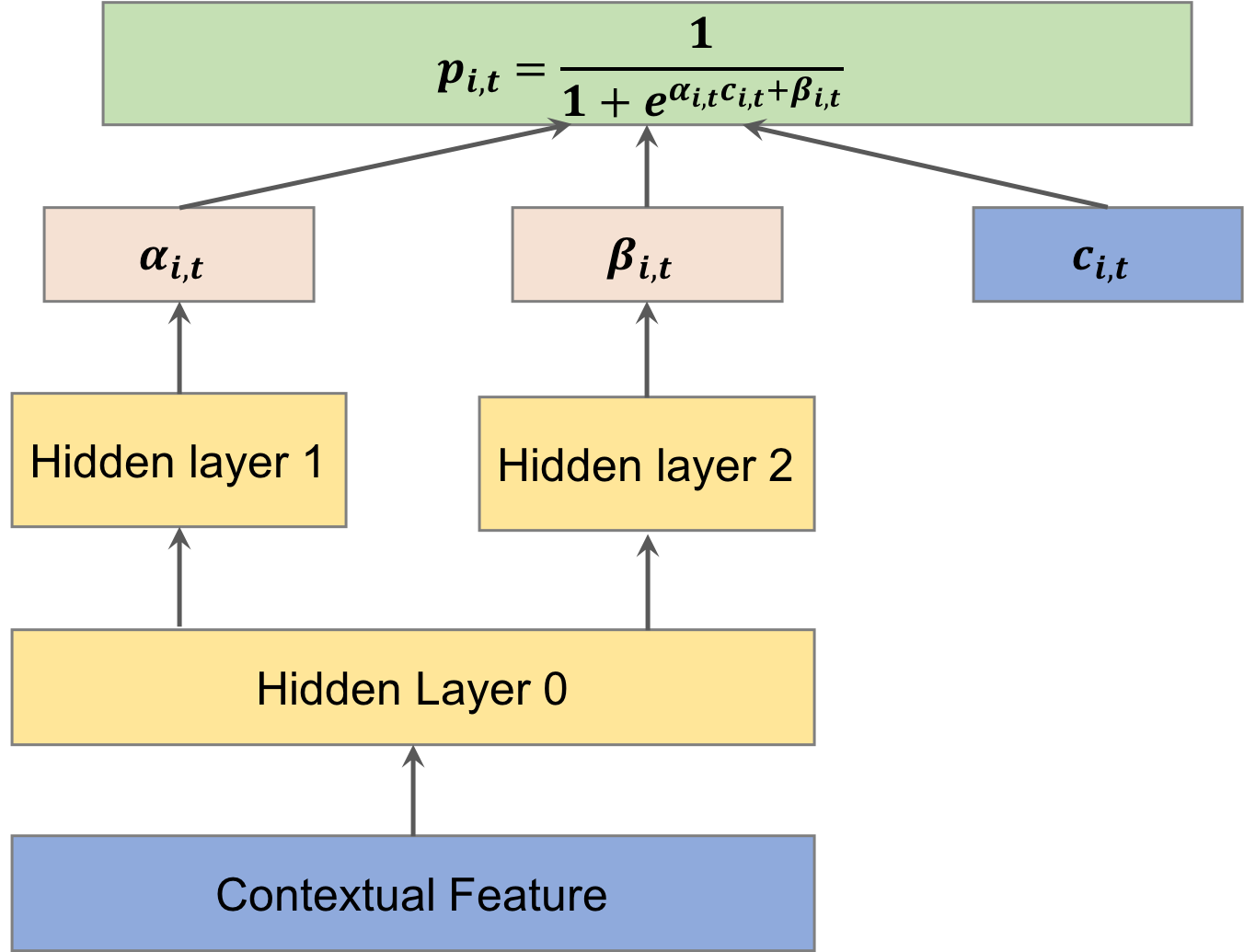}
  \caption{The illustration of the acceptance probability model.}\label{fig:network}
  \Description{need to add description}
\end{figure}

The training set $\{[\boldsymbol{x_{i,t}},c_{i,t}],p_{i,t}^*\}$ is constructed for each order $i$ at each allocation stage $t$ from historical observations. Contrary to the method in the reference\cite{zhao2019unified} where $\alpha_{i,t}$ and $\beta_{i,t}$ are learned separately, in this study, we learn $\alpha_{i,t}$ and $\beta_{i,t}$ simultaneously but with different hidden layers (see Figure \ref{fig:network}). In practice, the bonus is only allocated to a minority of orders such that the sample distribution of the training set is uneven. Therefore, we divide the training set into two kinds of batches: bonus batches with $c_{i,t}>0$ and normal batches with $c_{i,t}=0$. As in reference\cite{zhao2019unified}, $\beta_{i,t}$ is determined by contextual features $\boldsymbol{x_{i,t}}$. Therefore, to improve the performance of the model, the hidden layers of $\alpha_{i,t}$ and $\beta_{i,t}$ are updated using different kinds of batches. More specifically, the bonus batches are used to update the parameters of hidden layer 0 and layer 1, and the normal batches are used to do the same for hidden layer 0 and layer 2. Note that the attained $\alpha_{i,t}$ should be less than 0, which is consistent with the common sense of the more the bonus, the larger the acceptance probability.

\begin{figure}
  \centering
  \includegraphics[width=0.30\textwidth]{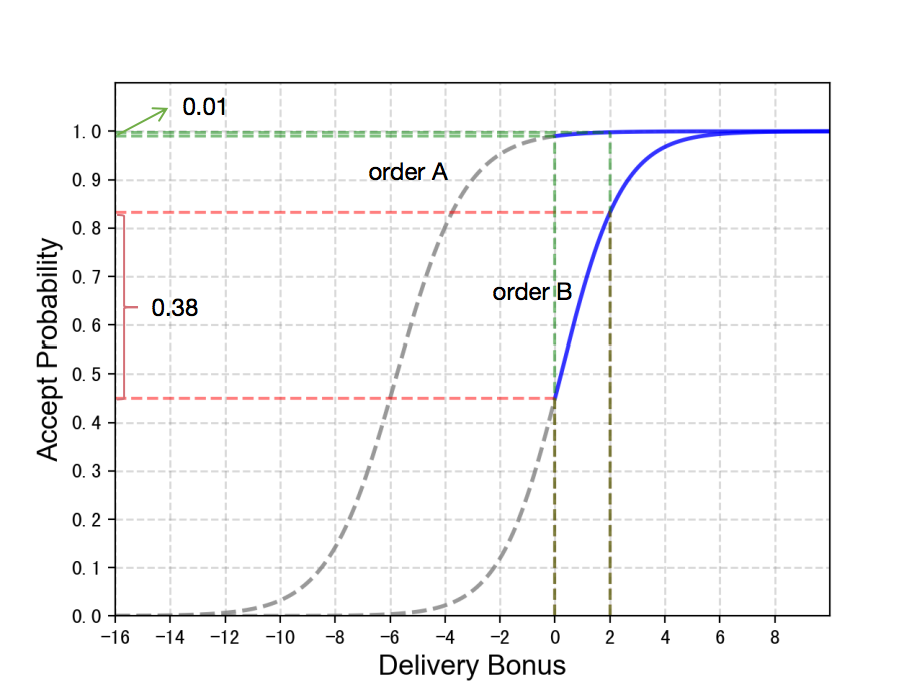}
  \caption{The relationship between the bonus and acceptance probability.}\label{fig:two_orders}
  \Description{need to add description}
\end{figure}

The acceptance probability model of orders is illustrated in Figure \ref{fig:two_orders}. As shown, the acceptance probabilities of the two orders may differ, even though their delivery bonus is the same. The acceptance probability (over 95\%) is much higher than that of order B(approximately 45\%) when the bonus is 0. Given the bonus of 2 RMB, the increment of acceptance probability of order A is 0.01, whereas order B is 0.38. Therefore, the motivation is to increase the total acceptance probability by allocating a bonus to order B.

The cancellation rate of order in $|T|$  stage also affects our decision, so we need to forecast it, which is marked as Symbol $q_{|T|}$. We have selected all orders entering the $|T|$  stage as a complete sample set. The characteristics of the sample are the relevant attributes of the order entering the stage, such as distance, environment, weather, and other characteristics. The label of the sample is that the canceled orders at this stage are marked as 1, and the other orders are marked as 0. We use the classic XGBoost model to train these samples to get the corresponding estimation model. Then we decompose the predicted value of the model at this stage. Each 0.05 forecast value interval is divided into one category, and each type of order is frequently sampled, and the proportion of positive samples within the interval is counted. This ratio is considered to be a predicted value which is the actual cancellation probability of the order in the region.

\subsection{Definition and Mathematical Model}\label{sec:model}
The set $N$ consists of the order placed within a week/month.The set $T$ means the bonus allocation stage set, i.e., the bonus allocated to an order can be changed at most $|T|$ times.Let $p_{i,t}$ be the acceptance probability of the order $i$ at allocation stage $t\in T$,$q_{i,t}$ be the cancellation probability of the order $i$ at allocation stage $t\in T$. The decision variable $c_{i,t}$ is the delivery bonus of the order $i\in N$ at allocation stage $t\in T$. The upper bound of the delivery bonus is $C_i^u$ . The function $p_{i,t}(c_{i,t})$ denotes the acceptance probability model of the order $i\in N$ if it is transited to allocation stage $t\in T$. The total delivery bonus cost should be within the given budget $B$.
The MSBA is modeled as follows:

\begin{align}
    \label{eq1.1}\max~& \sum_{i\in N}\sum_{t\in T}(\prod_{t_0=1}^{t-1} (1-p_{i,t_0}(c_{i,t_0})-q_{i,t_0}))p_{i,t}(c_{i,t}) \\
    \label{eq1.2}\mbox{s.t. }& \sum_{i\in N}\sum_{t\in T}(\prod_{t_0=1}^{t-1} (1-p_{i,t_0}(c_{i,t_0})-q_{i,t_0}))p_{i,t}(c_{i,t})c_{i,t}\le B\\
    \label{eq1.3}& 0\le c_{i,t}\le C_i^u \qquad \forall i \in N, t \in T
\end{align}

The objective function \eqref{eq1.1} maximizes the total acceptance probability, i.e., the excepted value of the accepted order quantity. Constraint \eqref{eq1.2} indicates that the expected value of the total delivery bonus cost should be within the given budget B. Constraints \eqref{eq1.3} restrain the delivery bonus $c_{i,t}$ within the given upper bound.

\begin{figure}
  \centering
  \includegraphics[width=0.4\textwidth]{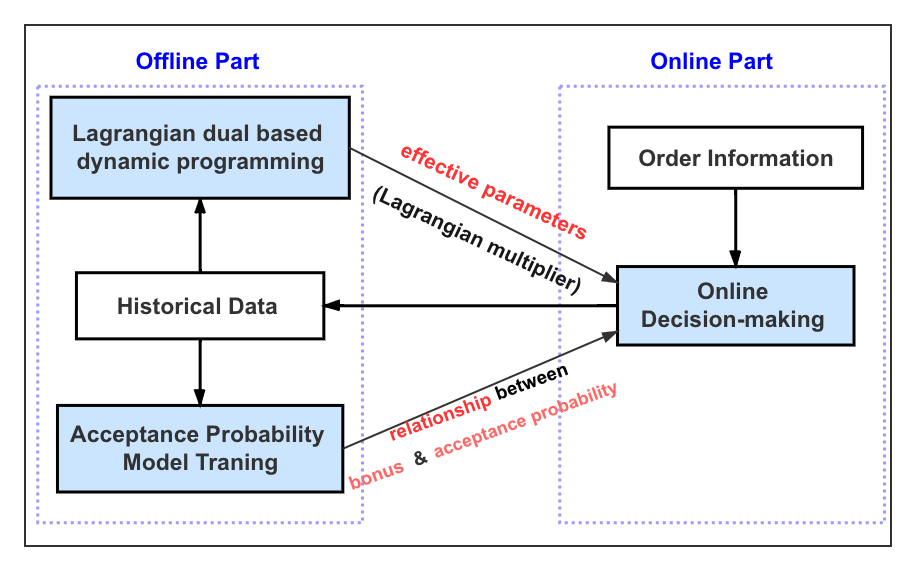}
  \caption{Our multi-stage bonus allocation framework.}\label{fig:framework}
  \Description{need to add description}
\end{figure}

\section{Algorithm}\label{sec:alg}
The framework of our bonus allocation approach is shown in Fig.~\ref{fig:framework}. Our algorithm contains an offline part, the Lagrangian Dual-based Dynamic Programming (LDDP), to solve Problem (3) based on the historical dataset, and an online part to satisfy fast real-time calculation of budget allocation. The offline part first optimizes Problem (3) and calculates the empirical Lagrangian multiplier $\lambda^{*}_t$ for each allocation stage $t$, and the online part uses the optimal $\lambda^{*}_t$ to make real-time bonus allocation decisions online.

\subsection{Offline Optimization: The Lagrangian Dual-based Dynamic Programming}
Problem (3) is a non-linear and non-convex multi-stage optimization problem, which is intractable in practice. In this section, we propose a novel LDDP method to solve this problem. We view this problem as a combination of two sub-problems. First, we allocate the total budget to each stage, and then within each stage, we calculate the optimal bonus for each order. The former sub-problem could be dealt with dynamic programming, and the latter one is a standard single-stage allocation problem solved by Lagrangian dual theory.

Let $N_{\tilde{t}}$ be the remaining order set transited to allocation stage $\tilde{t}$, if there is no budget posed on stage 1 to $\tilde{t}-1$. Thinking of the right-hand side of constraint \eqref{eq1.2} $\tilde{B}$ taking value $0, 1, 2, \cdots, B$ as \textit{money level}. Note that we discretize the budget to decrease the complexity of the algorithm and meet real-world conditions. The allocation stage subset $\{\tilde{t},\tilde{t}+1,...,|T|\}$ represented by $\tilde{t}$ as the "stage"(Note that this "stage" is used to define a sub-problem of dynamic programming, which is different from the allocation stage mentioned before), leads us to define the subproblem $P_{\tilde{t}}(\tilde{B})$ and the optimal value function $G_{\tilde{t}}(\tilde{B})$ as follows:
\begin{align}
    \label{eq2.1}G_{\tilde{t}}(\tilde{B})= \max~& \sum_{i\in N_{\tilde{t}}}\sum_{t=\tilde{t}}^{|T|}(\prod_{t_0=\tilde{t}}^{t-1} (1-p_{i,t_0}(c_{i,t_0})-q_{i,t_0}))p_{i,t}(c_{i,t}) \\
    \label{eq2.2}\mbox{s.t. }& \sum_{i\in N_{\tilde{t}}}\sum_{t=\tilde{t}}^{|T|}(\prod_{t_0=\tilde{t}}^{t-1} (1-p_{i,t_0}(c_{i,t_0})-q_{i,t_0}))p_{i,t}(c_{i,t})c_{i,t}\le \tilde{B}\\
    \label{eq2.3}& 0\le c_{i,t}\le C_i^u  ,\forall i \in N_{\tilde{t}}, t=\tilde{t},\tilde{t}+1,...,|T|
\end{align}
  Then, $z=G_{1}(B)$ gives us the optimal value of the MSBA\cite{wolsey1998integer}. 
  
  To define a recursion that allows us to calculate $G_{\tilde{t}}(\tilde{B})$ in terms of $G_{s}(\tilde{B_0})$ for $s>\tilde{t}$ and $\tilde{B_0} \le \tilde{B}$, we need to define a single-stage bonus allocation problem with the objective value function $g_{\tilde{t}}(\tilde{B})$, where budget $\tilde{B}$ is only spent on a single stage $\tilde{t}$. It indicates the optimal expected value of the accepted order quantity if budget $\tilde{B}$ is spent on a single stage $\tilde{t}$. Function $g_{\tilde{t}}(\tilde{B})$ is expressed as follows:
\begin{align}
    \label{eq3.1}g_{\tilde{t}}(\tilde{B})= \max~& \sum_{i\in N_{\tilde{t}}}p_{i,\tilde{t}}(c_{i,\tilde{t}}) \\
    \label{eq3.2}\mbox{s.t. }& \sum_{i\in N_{\tilde{t}}}p_{i,\tilde{t}}(c_{i,\tilde{t}})c_{i,\tilde{t}}\le \tilde{B}\\
    \label{eq3.3}& 0\le c_{i,\tilde{t}}\le C_i^u & \forall i \in N_{\tilde{t}}
\end{align}
By solving the single-stage bonus allocation problem with single-stage budget $\tilde{B} = k$, we can attain the accepted probability vector. $\boldsymbol{p}_{\tilde{t}}(k)=[p_{1,\tilde{t}}(c_{1,\tilde{t}}(k)),p_{2,\tilde{t}}(c_{2,\tilde{t}}(k)),...,p_{|N_{\tilde{t}}|,\tilde{t}}(c_{|N_{\tilde{t}}|,\tilde{t}}(k))]$, where $c_{i,\tilde{t}}(k)$ is the optimal bonus for order $i$  under total budget $k$ constraint on stage $\tilde{t}$.

 Then, $G_{\tilde{t}}(\tilde{B})$ can be expressed as follows:
\begin{equation}\label{eqas1.1}
\begin{aligned}
    G_{\tilde{t}}(\tilde{B})=\max_{k=0,1,...,\tilde{B}}\{g_{\tilde{t}}(k)+G'_{\tilde{t}+1}(\tilde{B}-k)\}
\end{aligned}
\end{equation}
 $G'_{\tilde{t}+1}(\tilde{B}-k)$ indicates the optimal value function with budget of $\tilde{B}-k$ spent on allocation stage set $\{\tilde{t}+1,...,|T|\}$, under the condition that budget of $k$ is spent on single allocation stage $\tilde{t}$. $G'_{\tilde{t}+1}(\tilde{B}-k)$ can be formulated as follows (Let $x_i$ be $\sum_{t=\tilde{t}+1}^{|T|}(\prod_{t_0=\tilde{t}}^{t-1} (1-p_{i,t_0}(c_{i,t_0})-q_{i,t_0}))p_{i,t}(c_{i,t})$ in the following expressions for short.):

\begin{align}
    \label{eqas2.1} G'_{\tilde{t}+1}(\tilde{B}-k)=\max~& \sum_{i\in N_{\tilde{t}}}(1-p_{i,\tilde{t}}(k)-q_{i,\tilde{t}})x_i \\
    \label{eqas2.2}\mbox{s.t. }& \sum_{i\in N_{\tilde{t}}}(1-p_{i,\tilde{t}}(k)-q_{i,\tilde{t}})x_ic_{i,t}\le \tilde{B}-k\\
    \label{eqas2.3}& 0\le c_{i,t}\le C_i^u  ,\forall i \in N_{\tilde{t}}, t=\tilde{t}+1,...,|T|
\end{align}
\subsubsection{Dimensionality Reduction}
Noticing that the definition of $G$ and $G'$ in \eqref{eqas1.1} is different, which is hard for dynamic programming. Hence, we must establish link between two optimization problems.

The main difference for $G_{\tilde{t}+1}(\tilde{B})$ and $G'_{\tilde{t}+1}(\tilde{B})$ is that $G_{\tilde{t}+1}(\tilde{B})$ is calculated by $\boldsymbol{1}^T\boldsymbol{x}$, while $G'_{\tilde{t}+1}(\tilde{B})$ is calculated by $(\boldsymbol{1}-\boldsymbol{p}_{\tilde{t}}(k)-\boldsymbol{q}_{\tilde{t}})^T\boldsymbol{x}$, in both objective function and constraint. A simple way to reduce complexity is to project vector $\boldsymbol{p}_{\tilde{t}}(k)+\boldsymbol{q}_{\tilde{t}}$ to an m-dimensional space under a predefined projection matrix, \textit{i.e.} $\boldsymbol{p}_{\tilde{t}}(k)+\boldsymbol{q}_{\tilde{t}} \approx \boldsymbol{H} \boldsymbol{u}$, where $\boldsymbol{H}$ is a predefined $(|N_{\tilde{t}}| \times m)$-dimensional projection matrix. Under such decomposition~\cite{birge1985decomposition}, the objective function of $G'_{\tilde{t}+1}(\tilde{B})$ could be written as a form of $\boldsymbol{1}^T\boldsymbol{x} - \boldsymbol{u}^T\boldsymbol{H}^T\boldsymbol{x}$.

To satisfy the computational requirement of industry in practice, we reduce the number of dimension $m$ to 1 in our work. In this simplest way, we choose $\boldsymbol{H}$ as $[1, 1, \cdots, 1]$ and hence $u=\frac{\sum_{N_{\tilde{t}}}(p_{i,\tilde{t}}(k)+q_{\tilde{t}})}{|N_{\tilde{t}}|}$ representing for the average probability for orders accepted or cancelled on stage $\tilde{t}$. We immediately build up the relationship between $G'_{\tilde{t}+1}(\tilde{B})$ and $G_{\tilde{t}+1}(\tilde{B})$: (1) the equivalence of the objective function: $\boldsymbol{1}^T\boldsymbol{x} - u \boldsymbol{H}^T\boldsymbol{x} = (1-u) \boldsymbol{1}^T\boldsymbol{x}$. (2) the equivalence of the constraint:
\begin{equation}
    \sum_{i\in N_{\tilde{t}}}(1-p_{i,\tilde{t}}(k)-q_{i,\tilde{t}})x_ic_{i,t}\le \tilde{B}  \iff \sum_{i\in N_{\tilde{t}}}x_ic_{i,t}\le \frac{\tilde{B}}{1-u}
\end{equation}
In other words, there must be $G'_{\tilde{t}+1}(\tilde{B}) = (1-u)G_{\tilde{t}+1}(\frac{\tilde{B}}{1-u})$. By dimensionality reduction, we sacrifice a little accuracy in change of low time complexity for dynamic programming~\cite{si2004handbook}.

\subsubsection{Recursion to Dynamic Programming}
 Formally, as $\sum_{i\in N_{\tilde{t}}}p_{i,\tilde{t}}(k)= g_t(k)$ holds according to the definition of $\boldsymbol{p}_{\tilde{t}}(k)$. Moreover, we define $\sum_{i\in N_{\tilde{t}}}q_{i,\tilde{t}}= Q_t$. Therefore, $G'_{\tilde{t}+1}(\tilde{B}-k)$ is formulated as follows:
\begin{align}
    \label{eqas4.1}G'_{\tilde{t}+1}(\tilde{B}-k)
    =\max~& \frac{|N_{\tilde{t}}|-g_t(k)-Q_t}{|N_{\tilde{t}}|} \sum_{i \in N_{\tilde{t}}}x_i \\
    \label{eqas4.2}\mbox{s.t. }& \sum_{i\in N_{\tilde{t}}}x_ic_{i,t}\le \frac{|N_{\tilde{t}}|}{|N_{\tilde{t}}|-g_t(k)-Q_t}\tilde{B}-k\\
    \label{eqas4.3}& 0\le c_{i,t}\le C_i^u  ,\forall i \in N_{\tilde{t}}, t=\tilde{t}+1,...,|T|
\end{align}

In summary, the expression \eqref{eqas1.1} can be reformulated as the recursion as follows:
\begin{equation}\label{eqr1.1}
\begin{aligned}
    G_{\tilde{t}}(\tilde{B})=\max_{k=0,1,...,\tilde{B}}\{g_{\tilde{t}}(k)+\frac{|N_{\tilde{t}+1}|}{|N_{\tilde{t}}|}* G_{\tilde{t}+1}[(\tilde{B}-k)\frac{|N_{\tilde{t}}|}{|N_{\tilde{t}+1}|}]\}
\end{aligned}
\end{equation}
where $|N_{\tilde{t}+1}|=|N_{\tilde{t}}|-g_t(k)-Q_t$. Starting the recursion with $G_{|T|}(\tilde{B})=g_{|T|}(\tilde{B})$ for $\tilde{B} \ge 0$, we use the recursion \eqref{eqr1.1} to successively calculate $G_{|T|-1},G_{|T|-2},...,G_1$ for all integral values of $\tilde{B}$ from 0 to B. An overview of the dynamic programming algorithm is presented in Algorithm \ref{alg:DP}. In this process, we must solve a single-stage bonus allocation problem with objective value of $g_{\tilde{t}}(\tilde{B})$ for all $\tilde{t}\in T$ and $\tilde{B}\in \{0,1,2,...,B\}$. 

\begin{algorithm}[t]
\DontPrintSemicolon
 \SetKwComment{Comment}{$\triangleright$\ }{}
 \SetKw{KwBy}{by}
   \SetNoFillComment
    \caption{Lagrangian dual-based Dynamic Programming}\label{alg:DP}
        \KwIn{Allocation stage set $T$, order set $N_{\tilde{t}}, \forall t\in T$, total budget B} 
        \KwOut{empirical Lagrangian Multipliers $\lambda^*[|T|]$} 
        \BlankLine
        \BlankLine
        \Comment*[l]{Solve the single-stage bonus allocation repeatedly and obtain $g[|T|][B]$ and $\lambda[|T|][B]$}
        \ForAll{$\tilde{t}\in T$}{
        \While{$\tilde{B}$<=B}{
        $g[\tilde{t}][\tilde{B}],\lambda[\tilde{t}][\tilde{B}]\gets BA(N_{\tilde{t}},\tilde{B})$\;
        $\tilde{B}\gets \tilde{B}+1$\;
        }
        }
        \BlankLine
        \BlankLine
        \Comment*[l]{Solve the sub-problem $P_{\tilde{t}}(\tilde{B})$ recursively and obtain G[|T|][B], then estimate the probability of order cancellation at stage ${\tilde{t}}$ and sum it up,get the corresponding $Q[\tilde{t}]$ }
        $G[|T|] \gets g[|T|]$\;
        $\tilde{t}\gets |T|-1$\;
        $\tilde{B}\gets 0$\;
        \While{$\tilde{t}\ge 1$}{
        \While{$\tilde{B}$<=B}{
        $G[\tilde{t}][\tilde{B}]\gets G[\tilde{t}+1][\tilde{B}]$\;
        $k^*\gets 0$\;
        \For{$k\gets0$ \KwTo $\tilde{B}$ \KwBy $1$}{
        $N'\gets |N_{\tilde{t}}|-g[\tilde{t}][k] - Q[\tilde{t}]$\;
        $temp\gets g[\tilde{t}][k]+\frac{N'}{|N_{\tilde{t}}|}G[\tilde{t}+1][\lfloor{\frac{|N_{\tilde{t}}|}{N'}(\tilde{B}-k)}\rfloor]$\;
        \If{$G[\tilde{t}][\tilde{B}]< temp$}{$G[\tilde{t}][\tilde{B}]\gets temp$\;
        $k^*\gets k$\;
        }
        }
        $a[\tilde{t}][\tilde{B}]\gets k^*$\Comment*[l]{a[][] is used to backtrack the optimal Lagrangian multiplier for each allocation stage}
        $\tilde{B}\gets \tilde{B}+1$\;
        }
        $\tilde{t}\gets \tilde{t}-1$
        }
       \Comment*[l]{Backtrack the optimal solution and record the optimal empirical Lagrangian multiplier for each stage into arrays $\lambda^*[|T|]$}
       $B_0 \gets B$\;
       \For{$t=1$ \KwTo $|T|$ \KwBy $1$}{
      $B^* \gets a[t][B_0]$\;
      $\lambda^*[t] \gets \lambda[t][B^*]$\; 
      $B_0 \gets B_0 -B^*$
       }
        return $\lambda^*[|T|]$
\end{algorithm}

\subsubsection{Solve Single-stage Bonus Allocation Problem Based on Lagrangian Dual Theory}\label{sec:offline-single}
As defined by functions \eqref{eq3.1}-\eqref{eq3.3}, both the objective function and constraint functions are non-convex with respect to $\boldsymbol{c}$. Therefore, solving the problem directly is very difficult. Inspired by \cite{zhao2019unified,dong2009dynamic,li2011pricing}, we reformulate the problem into an equivalent convex optimization problem. Because $p_{i,t}(c_{i,t})=\frac{1}{1+e^{\alpha_{i,t} c_{i,t}+\beta_{i,t}}}$, we have
\begin{equation}\label{eq4.1}
\begin{split}
     & c_{i,t}(p_{i,t})=-\frac{\beta_{i,t}}{\alpha_{i,t}}+\frac{1}{\alpha_{i,t}}(\ln(1-p_{i,t})-\ln(p_{i,t}))\\
     & p_{i,t}\in(0,1), \forall i\in N,t\in T
\end{split}
\end{equation}

Consequently, the constraint function \eqref{eq3.2} is rewritten as a function of $p_{i,\tilde{t}}$:
\begin{equation}
    f_{\tilde{t}}(\boldsymbol{p})=\sum_{i\in N_{\tilde{t}}}p_{i,\tilde{t}}c_{i,\tilde{t}}(p_{i,\tilde{t}})\\
\end{equation}

Referring to \cite{zhao2019unified,dong2009dynamic,li2011pricing}, $f_{\tilde{t}}(\boldsymbol{p})$ is a convex function in $\boldsymbol{p}$ for $\forall \tilde{t}\in T$. 

Therefore, the original problem \eqref{eq3.1}-\eqref{eq3.3} is reformulated as follows:
\begin{align}
    \label{eq5.1}-g_{\tilde{t}}(\tilde{B})= \min~& -\sum_{i\in N_{\tilde{t}}}p_{i,\tilde{t}} \\
    \label{eq5.2}\mbox{s.t. }& \sum_{i\in N_{\tilde{t}}}p_{i,\tilde{t}}c_{i,\tilde{t}}(p_{i,\tilde{t}})\le \tilde{B}\\
    \label{eq5.3}& P_i^l\le p_{i,\tilde{t}}\le P_i^u & \forall i \in N_{\tilde{t}}
\end{align}
Clearly, this is a differentiable convex optimization problem with respect to $\boldsymbol{p}$ and Slater's condition holds easily. The duality gap between the primal and dual objective value equals to 0\cite{boyd2004convex}.

By introducing a dual variable $\lambda$, the Lagrangian relaxation function of the single-stage bonus allocation problem is represented as follows:
\begin{equation}
   \label{eqs1.1} L(\boldsymbol{p},\lambda)= ~-\sum_{i\in N_{\tilde{t}}} p_{i,\tilde{t}}+\lambda [\sum_{i\in N_{\tilde{t}}}p_{i,\tilde{t}}c_{i,\tilde{t}}(p_{i,\tilde{t}})- \tilde{B}]
\end{equation}
where $P_i^l\le p_{i,\tilde{t}}\le P_i^u $,$\forall i\in N, \lambda \in \mathbb R_+$.
Correspondingly, the Lagrangian dual problem is formulated as:
\begin{equation}
    \label{eqs1.2}\max_{\lambda \in \mathbb R_+}\min_{P_i^l\le p_{i,\tilde{t}}\le P_i^u} L(\boldsymbol{p},\lambda)
\end{equation}

As mentioned before, the primal optimal value $-g_{\tilde{t}}(\tilde{B})$ is equal to the dual optimal value owing to the zero duality gap. By introducing a bisection algorithm represented in Algorithm \ref{alg0}, we can obtain the optimal solution $\lambda_{\tilde{t}}(\tilde{B})$ and the optimal value $g_{\tilde{t}}(\tilde{B})$.

\begin{algorithm}[t]
\DontPrintSemicolon
\SetKwComment{Comment}{$\triangleright$\ }{}
 \SetKw{KwBy}{by}
   \SetNoFillComment
    \caption{Bisection Algorithm for a single-stage problem(BA($N_{\tilde{t}}$,$\tilde{B}$))}\label{alg0}
        \KwIn{Order set $N_{\tilde{t}}$,  budget $\tilde{B}$} 
        \KwOut{$\lambda_t(\tilde{B})$ and $g_{\tilde{t}}(\tilde{B})$} 
        $low \gets 0$ \;
        $high \gets M$\Comment*[l]{M is a big number}
        \While{$high-low> \epsilon$}{
        $s\gets 0$\;
        $opt\gets 0$\;
        $mid \gets \frac{high+low}{2}$\;
        \ForAll{$i\in N_{\tilde{t}}$}{
        $p_{i,\tilde{t}}^* \gets \arg \min_{P_i^l\le p_{i,\tilde{t}}\le P_i^u}-p_{i,\tilde{t}}+mid*p_{i,\tilde{t}}c_{i,\tilde{t}}(p_{i,\tilde{t}})$\;
        $s\gets s+ p_{i,\tilde{t}}^*c_{i,\tilde{t}}(p_{i,\tilde{t}}^*)$\;
        $opt\gets opt- p_{i,\tilde{t}}^*$\;
        }
        \If{$s-\tilde{B}\ge 0$}{$low \gets mid $}
        \Else{$high \gets mid$}
        }
        return $\lambda_t(\tilde{B}) \gets high$, $g_{\tilde{t}}(\tilde{B})\gets -opt$
\end{algorithm}

\subsection{Online Allocation Algorithm}
Given the multiplier $\lambda^*_{\tilde{t}}$ attained by the LDDP algorithm for stage $\tilde{t}\in T$, the online problem is written as:
\begin{equation}
   \label{eqo1.1} \min_{0\le c_{i,\tilde{t}}\le C_i^u} ~\{-\sum_{i\in N_{\tilde{t}}} p_{i,\tilde{t}}(c_{i,\tilde{t}})+\lambda^*_{\tilde{t}} [\sum_{i\in N_{\tilde{t}}}p_{i,\tilde{t}}(c_{i,\tilde{t}})c_{i,\tilde{t}}]\}
\end{equation}
which is equivalent to:
\begin{equation}
   \label{eqo1.2} \min_{0\le c_{i,\tilde{t}}\le C_i^u} ~\{\sum_{i\in N_{\tilde{t}}} (\lambda^*_{\tilde{t}}p_{i,\tilde{t}}(c_{i,\tilde{t}})c_{i,\tilde{t}}-p_{i,\tilde{t}}(c_{i,\tilde{t}}))\}
\end{equation}

The above problem is reformulated as the summation of |$N_{\tilde{t}}$| separable minimizing problems as follows:
\begin{equation}
   \label{eqo1.3} \sum_{i\in N_{\tilde{t}}} \min_{0\le c_{i,\tilde{t}}\le C_i^u}\{ (\lambda^*_{\tilde{t}}p_{i,\tilde{t}}(c_{i,\tilde{t}})c_{i,\tilde{t}}-p_{i,\tilde{t}}(c_{i,\tilde{t}}))\}
\end{equation}
Therefore, to determine the delivery bonus for a specific order, only a one-dimensional problem needs to be solved. An optimal solution for each separated problem can easily be obtained online. The optimal delivery bonus of a specific order $i\in N$ is expressed as follows:
\begin{equation}
\begin{aligned}
    \label{eqo1.4}c^*_{i,\tilde{t}}&= \arg \min_{0\le c_{i,\tilde{t}}\le C_i^u} \lambda^*_{\tilde{t}}p_{i,\tilde{t}}(c_{i,\tilde{t}})c_{i,\tilde{t}}-p_{i,\tilde{t}}(c_{i,\tilde{t}})
\end{aligned}
\end{equation}
Because the bonus of the order $i$ ranges from 0 to $C_i^u$, and the number of potential bonuses is limited, the optimization problem \eqref{eqo1.4} is easily solved by enumerating the potential bonus. Other numerical optimization methods are applicable, but we are not going to investigate them further here. Note that multiple problems can be solved in parallel here because the first problem is separated into multiple independent problems. The bonus decision for each order can be traversed to obtain the optimal solution. The complexity of determining the delivery bonus of an order online is O(1).

\subsection{Periodic Control}
At the beginning of the month, the total budget is worked out by business analysts, but the actual future monthly order size and distribution are unknown. Fortunately, history data shows that the monthly order distribution of the meal delivery platform is relatively stable and the deviation between offline historical orders and online orders is slight. To further satisfy the budget constraint under uncertain online scenario, we adopt some simple but effective periodic control strategies.

The first strategy is that we execute the offline decision-making system once a day, and the selected training set is the order data of the previous month from the current day. The target budget for offline training is calculated by the remaining budget of this month divided by the expected future order numbers of this month and multiply the total order numbers in the past 30 days. The second strategy is 
to adopt some simple control methods on realtime expenditure ~\cite{hao2015soft}. For example, when the proportion of total realtime expenditure to total budget is greater than 110\%, we will reduce the bonus to some proportion ratio. When the proportion is less than 90\%, we will increase the bonus. The ratio of upward and downward adjustment is positively correlated with the difference between the total realtime expenditure and total budget. These two strategies ensure that the online realtime expenditure could be controlled around the pre-defined budget within an acceptable range.

\section{Experiments}\label{sec:exp}

To verify the effectiveness of this method, we present the results of both offline experiments and online A/B tests.

\subsection{Performance of offline experiments}

In this section, we conducted several offline experiments. The instances are generated on real-world data derived from the Meituan meal delivery platform. Without the loss of generality, we chose the instances of four cities named Lanzhou, Nanchang, Weihai, and Chengdu with different typical order sizes over one week. Specifically, the data sets contain 46,978, 71,427, 120,017, and 258,612 orders, respectively. The delivery bonus of each order is calculated within the upper bound, which the business manager sets. It is worth mentioning that the \textit{budget} in the figures below indicates the \textit{average budget} for each order, which makes the results comparable among instances with different numbers of orders. A typical order life cycle starts from the order placed time to order delivered time or order canceled time, and the number of allocation stages indicates the dynamic bonus allocation frequency.  For example, if the order life cycle is 50 minutes and the number of allocation stages is 10, the bonus allocation decision is determined every 5 minutes.

\subsubsection{Performance of the MSBA}

\begin{figure}
     \centering
     \begin{subfigure}[b]{0.225\textwidth}
         \centering
         \includegraphics[scale=0.25]{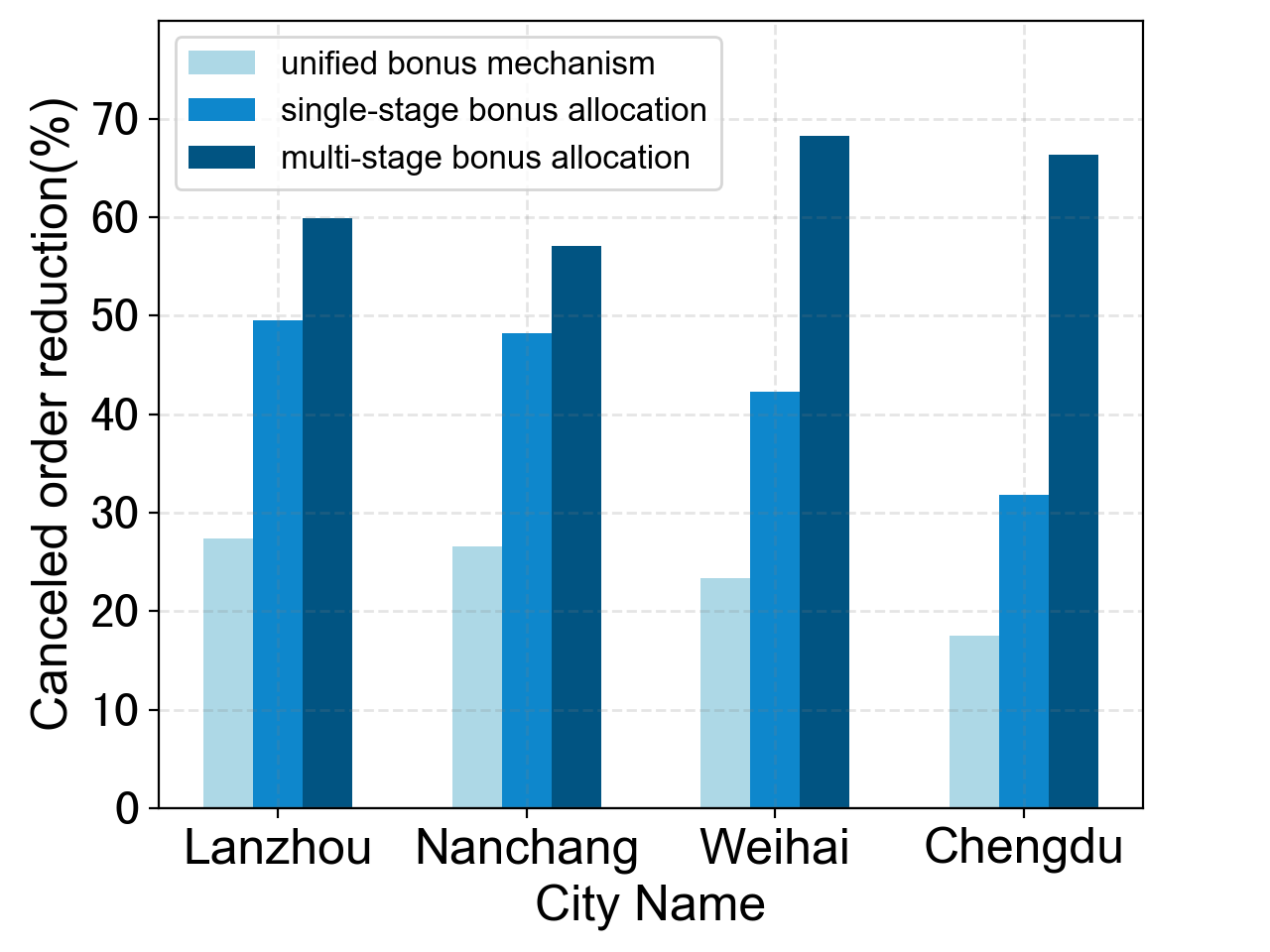}
         \caption{}
         \label{fig:m-pricing2}
     \end{subfigure}
     \begin{subfigure}[b]{0.225\textwidth}
         \centering
         \includegraphics[scale=0.25]{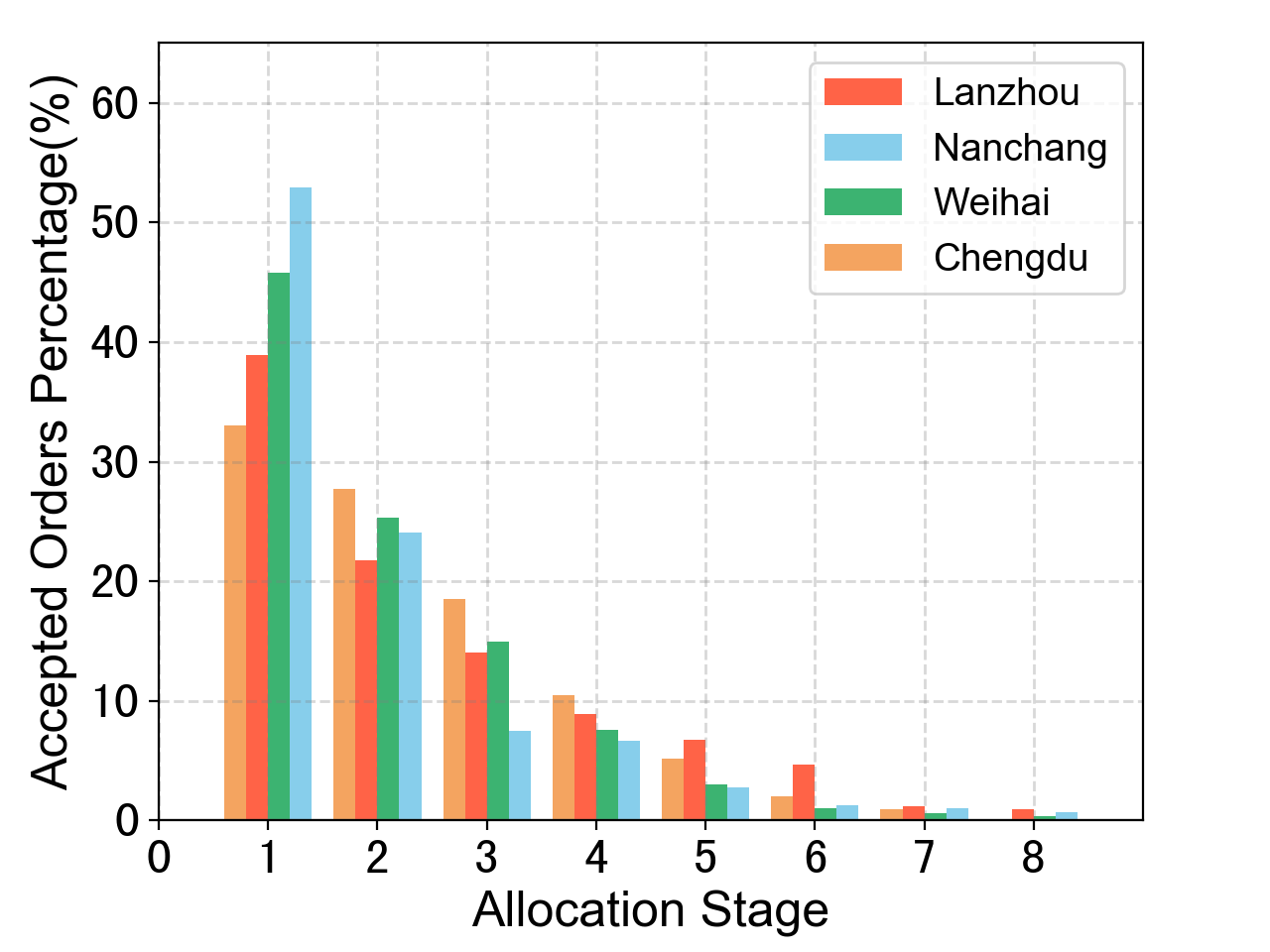}
         \caption{}
         \label{fig:m-pricing3}
     \end{subfigure}
     \hfill
        \label{fig:two graphs}
        \caption{(a) Comparison of the canceled order reduction with different approaches.  (b) The number of accepted orders on multiple allocation stages.}
\end{figure}

In this section, we examine the performance of the MSBA method considering eight allocation stages and about a six-minute time interval between two allocation stages. A comparison of the results with the unified bonus mechanism in which the same delivery bonus is allocated to each order, single-stage bonus allocation\cite{zhao2019unified} and MSBA is presented in Figure \ref{fig:m-pricing2}. As shown in the figure, compared to the cases without bonus allocation, using an MSBA reduces the number of canceled orders by more than 60\%. Moreover, it also shows that the MSBA outperforms the other two approaches. More specifically, compared to multi-stage bonus allocation and the single-stage bonus allocation, the number of canceled orders derived from the MSBA is around 20\% and 40\% lower, respectively. More importantly, for the instances with a more significant number of orders, the MSBA performs better, whereas the unified bonus mechanism and single-stage bonus allocation perform worse.

\begin{figure}[tb]
    \centering
    \begin{subfigure}[b]{0.22\textwidth}
    \includegraphics[width=\textwidth]{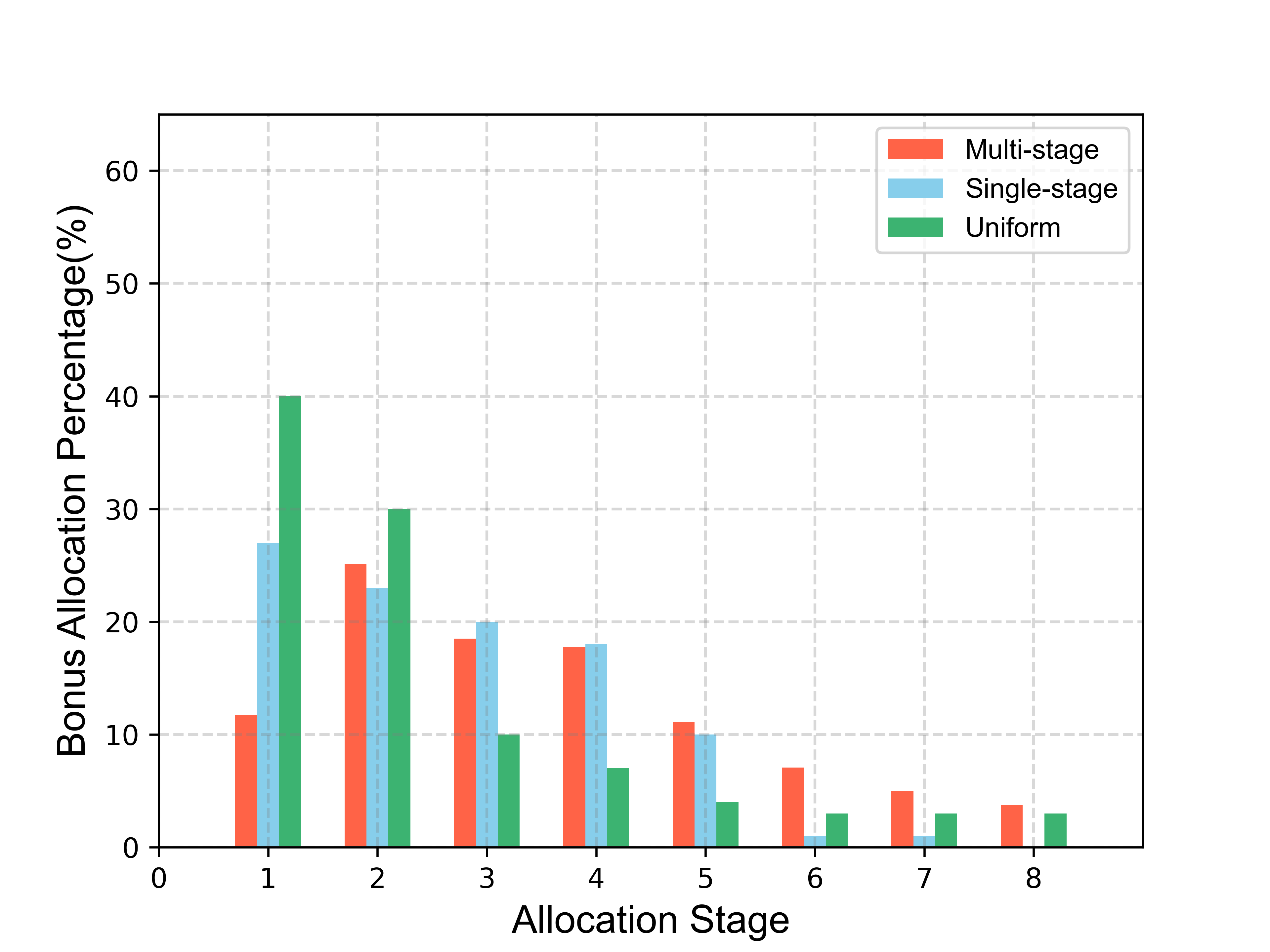}
    \end{subfigure}
    % \quad
    \begin{subfigure}[b]{0.22\textwidth}
    \includegraphics[width=\textwidth]{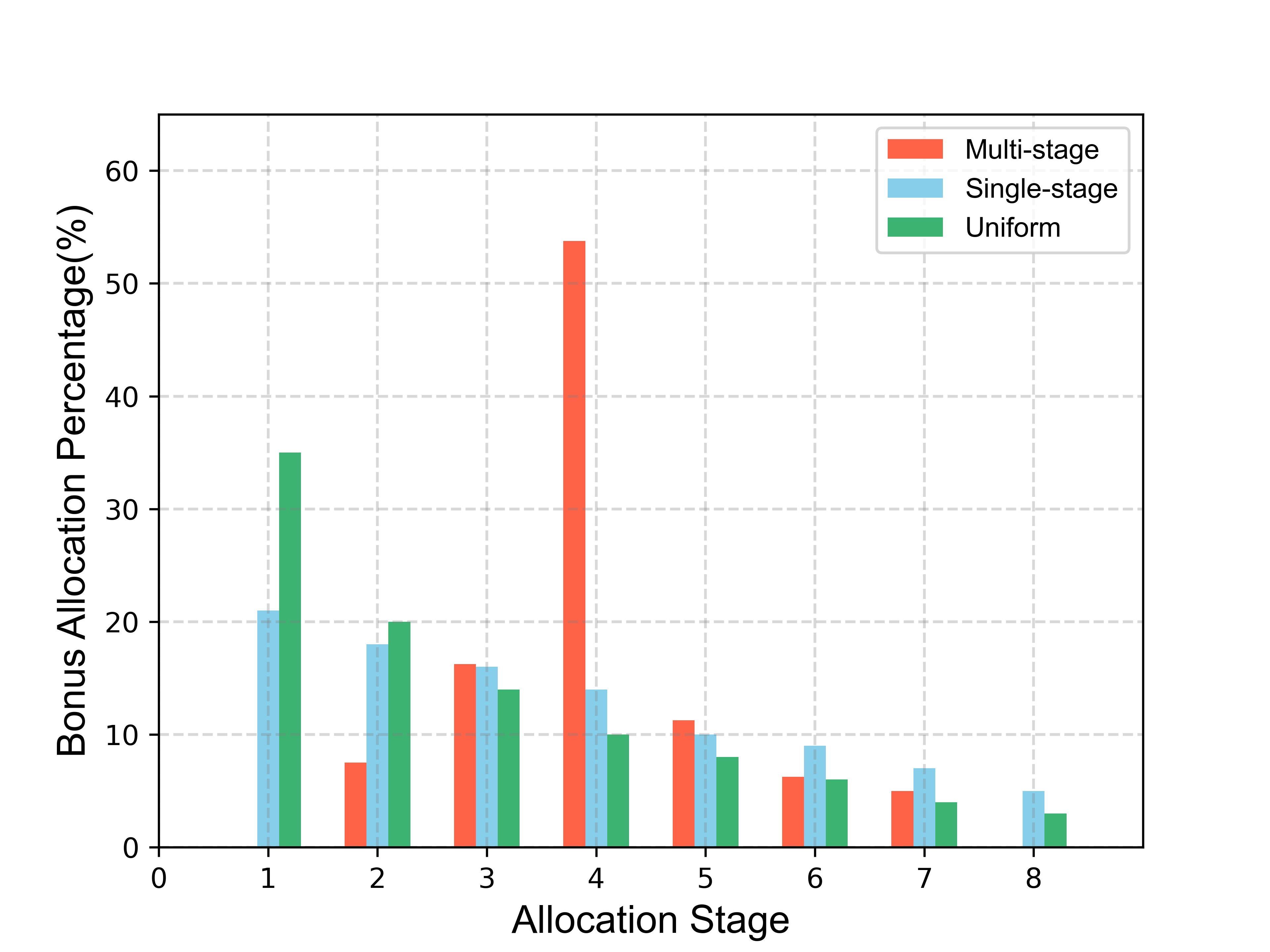}
 
    \end{subfigure}
    
    \begin{subfigure}[b]{0.22\textwidth}
    \includegraphics[width=\textwidth]{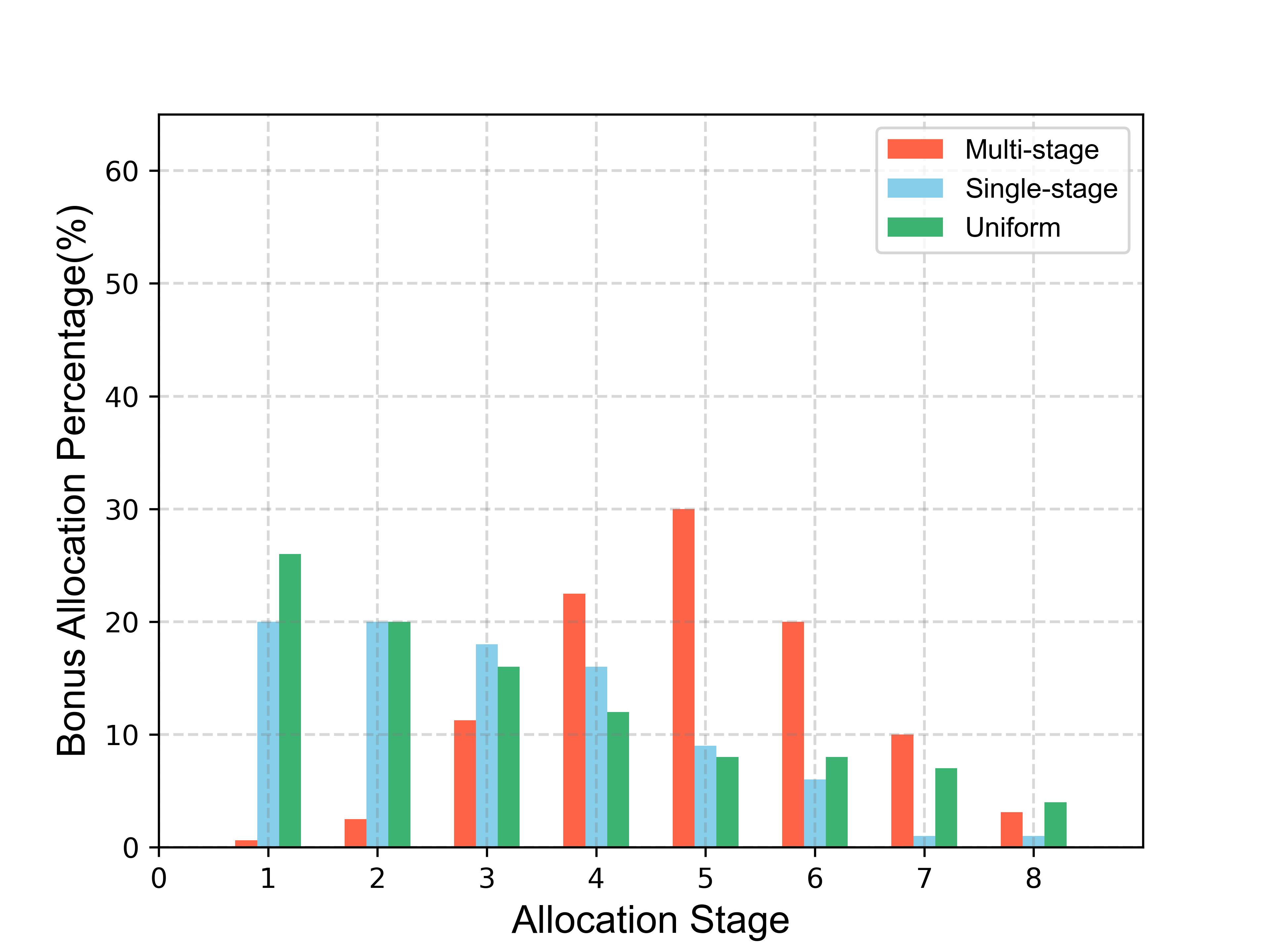}
 
    \end{subfigure}
    \begin{subfigure}[b]{0.22\textwidth}
    % \qquad
    \includegraphics[width=\textwidth]{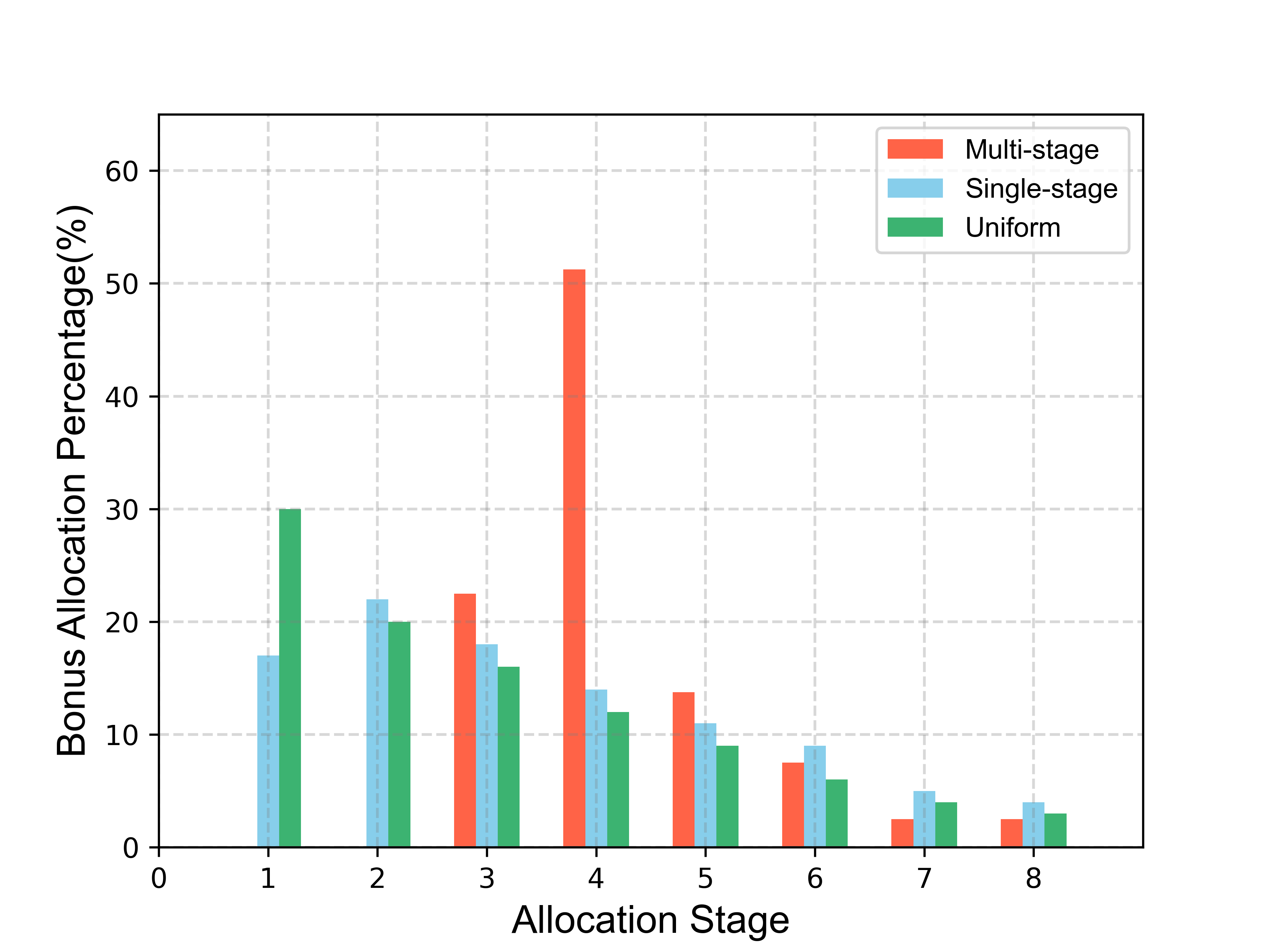}
 
    \end{subfigure}
    \caption{Total bonus allocation among all allocation stages with different approaches.}
    \label{fig:m-pricing}
\end{figure}

Figure \ref{fig:m-pricing} illustrates the total bonus allocation for each allocation stage for different methods. We observe that the total bonus allocation increases at first and then decreases as the allocation stage extends. To determine the potential reasons, we investigate the distribution of the accepted order numbers on these allocation stages. As shown in Figure \ref{fig:m-pricing3}, most orders are accepted immediately at the first allocation stage, and the number of accepted orders decreases as the allocation stage extends. Recall from Section \ref{sec:pro-model} that our motivation was to increase the total acceptance probabilities of all orders by allocating delivery bonus to "order B" instead of "order A" (Figure \ref{fig:two_orders}). Correspondingly, orders with high acceptance probability play the role of "order A". Consequently, the total bonus allocated to the first stage is small because most "order A" instances are accepted at the first stage, and the total allocated bonus increases at the early stages because of the number of "order B" increases. Then, as the allocation stage extends, the total bonus decreases since the remaining orders decrease.

% \begin{figure}
%     \centering
%     \includegraphics[width=0.3\textwidth]{m-pricing-2.png}
%     \caption{Comparison of the canceled order reduction with different approaches.}
%     \label{fig:m-pricing2}
% \end{figure}

% \begin{figure}
%     \centering
%     \includegraphics[width=0.3\textwidth]{m-pricing.png}
%     \caption{The number of accepted orders on multiple allocation stages.}
%     \label{fig:m-pricing3}
% \end{figure}

% \begin{figure}[tb]
%     \centering
%     \begin{subfigure}[b]{0.2\textwidth}
%     \includegraphics[width=\textwidth]{m-pricing2-a.png}

%     \end{subfigure}
%     % \quad
%     \begin{subfigure}[b]{0.2\textwidth}
%     \includegraphics[width=\textwidth]{m-pricing2-b.png}
 
%     \end{subfigure}
    
%     \begin{subfigure}[b]{0.2\textwidth}
%     \includegraphics[width=\textwidth]{m-pricing2-c.png}
  
%     \end{subfigure}
%     \begin{subfigure}[b]{0.2\textwidth}
%     % \qquad
%     \includegraphics[width=\textwidth]{m-pricing2-d.png}
 
%     \end{subfigure}
%     \caption{The comparison of the number of accepted orders between multistage dynamic bonus allocation and uniform bonus mechanism.}
%     \label{fig:m-pricing2}
% \end{figure}

\subsubsection{Performance of bonus allocation methods with different budget}

\begin{figure}[tb]
    \centering
    \begin{subfigure}[b]{0.22\textwidth}
    \includegraphics[width=\textwidth]{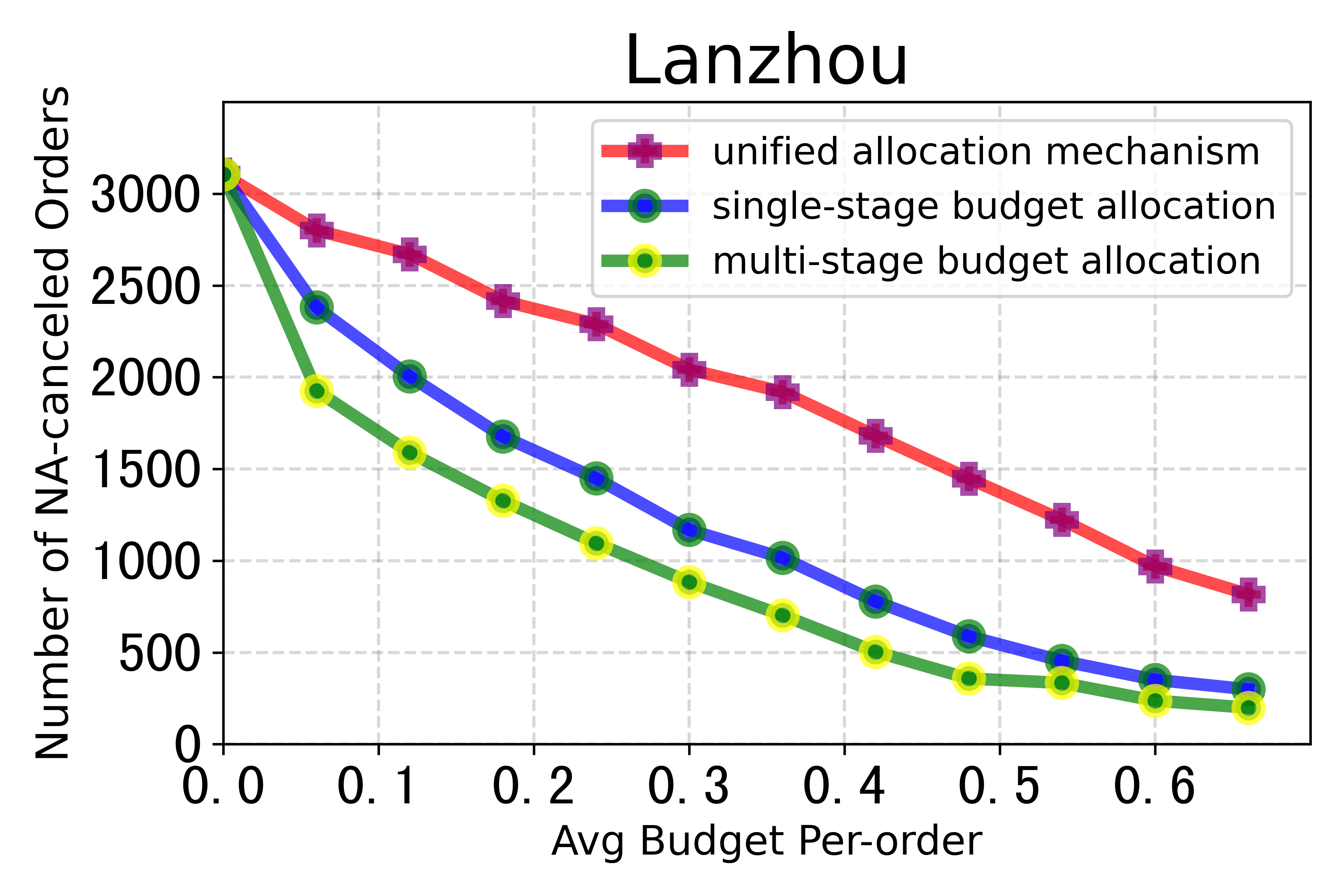}
    \end{subfigure}
    % \quad
    \begin{subfigure}[b]{0.22\textwidth}
    \includegraphics[width=\textwidth]{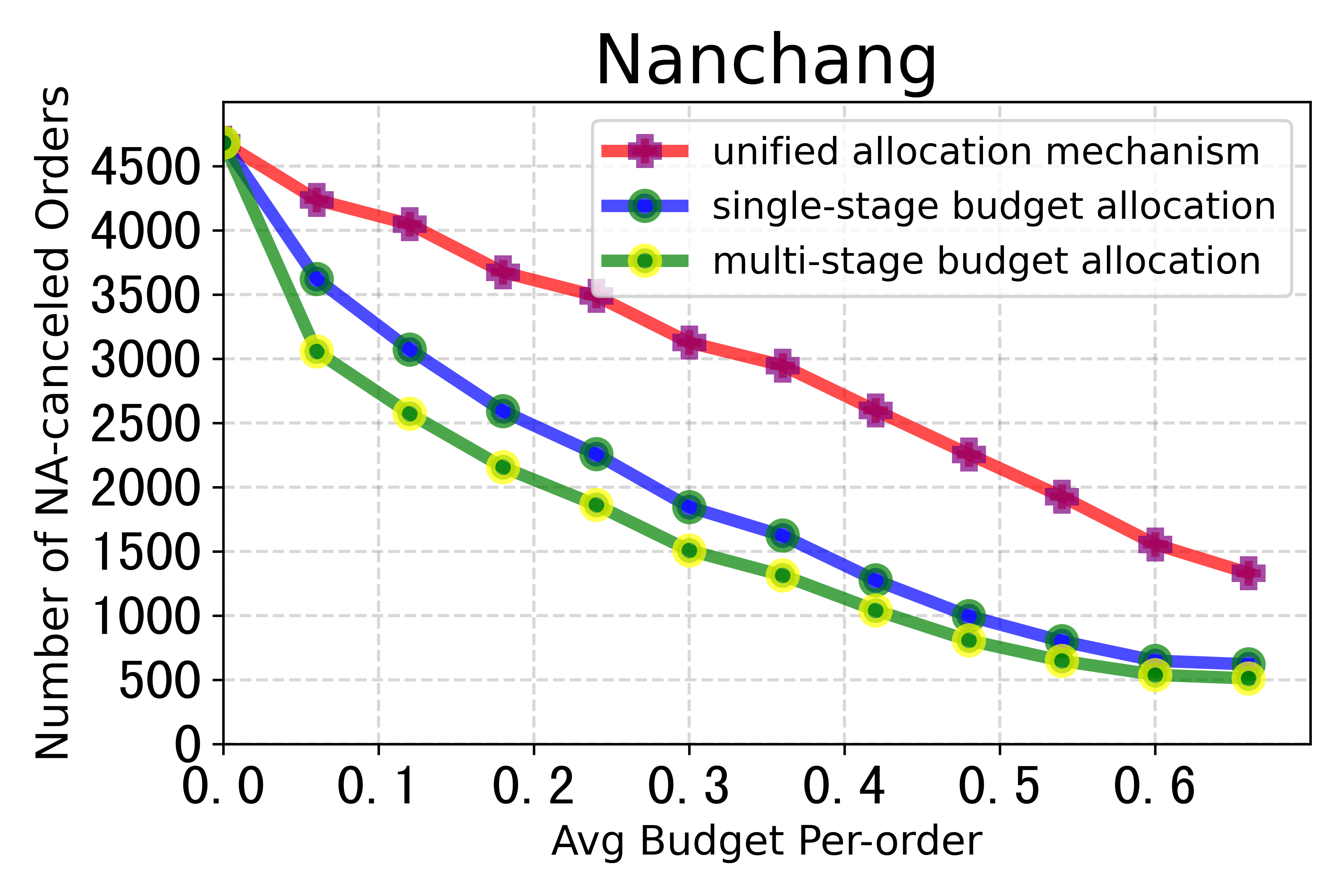}
 
    \end{subfigure}
    
    \begin{subfigure}[b]{0.22\textwidth}
    \includegraphics[width=\textwidth]{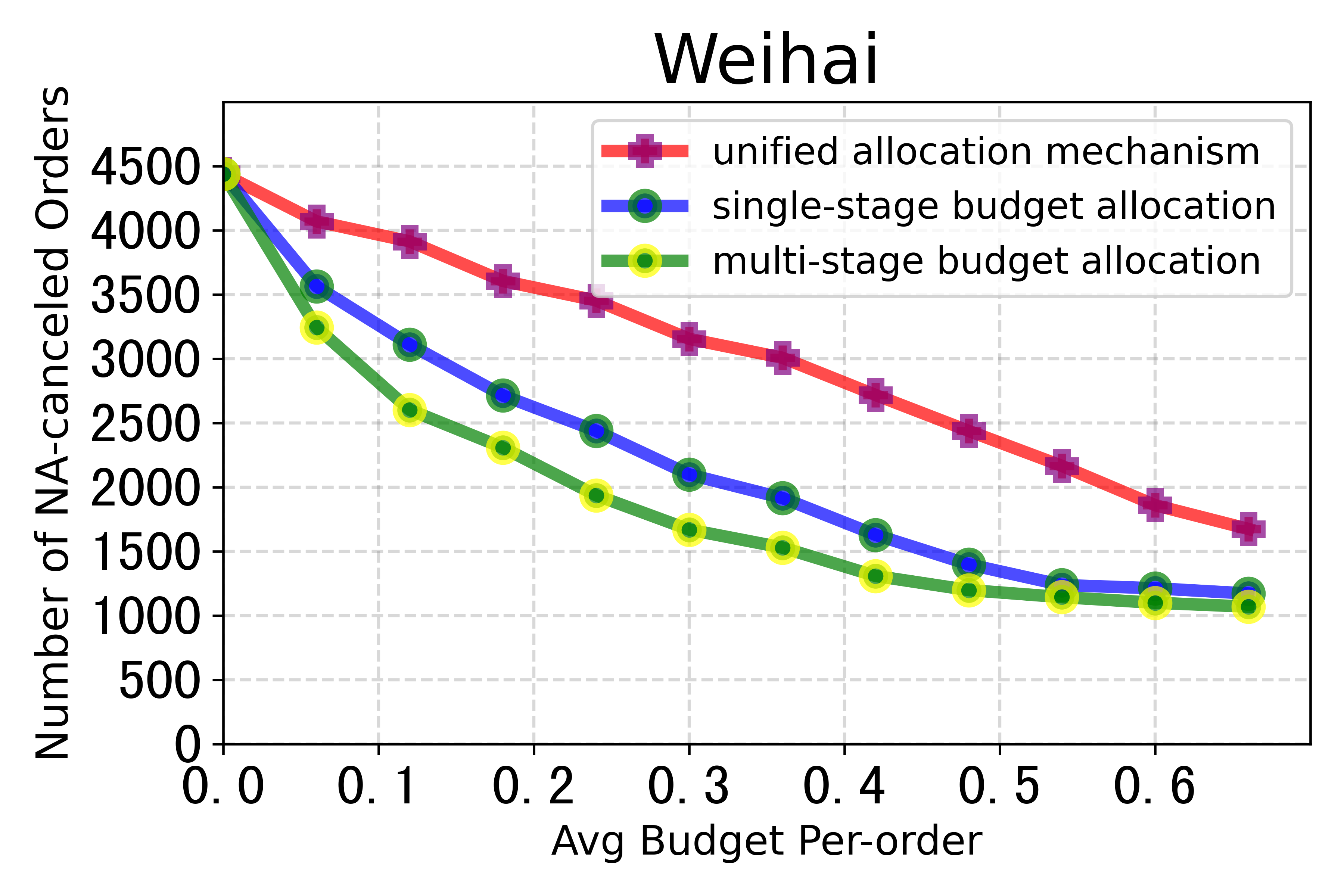}
 
    \end{subfigure}
    \begin{subfigure}[b]{0.22\textwidth}
    % \qquad
    \includegraphics[width=\textwidth]{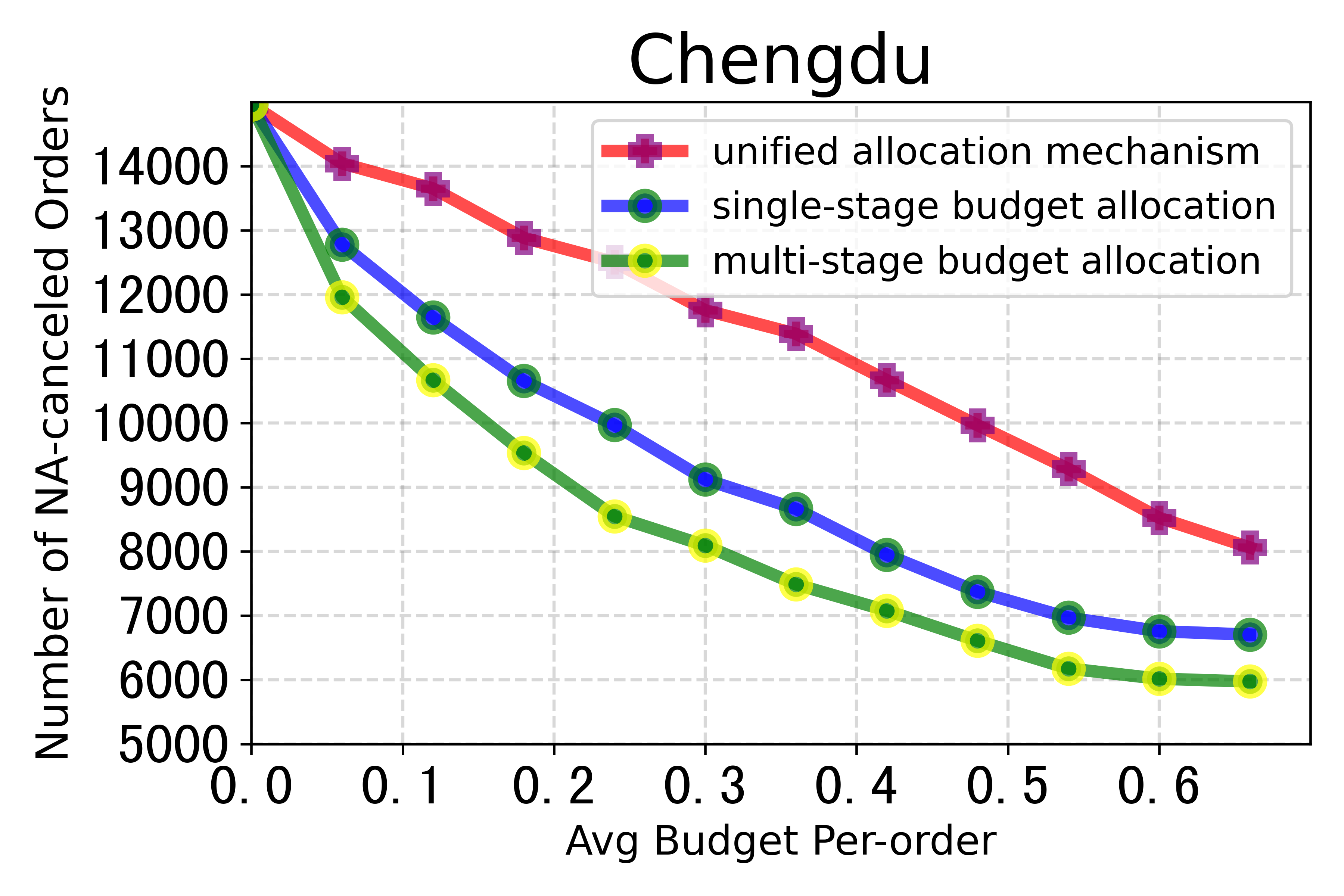}
 
    \end{subfigure}
     \caption{Comparison of the NA-canceled order reduction
with different approaches at diffent budget. }
    \label{fig:s-pricing}
\end{figure}

In this section, we analyze the performance of bonus allocation methods with different budgets. We compare the results derived by the MSBA ,the single-stage allocation and the unified bonus mechanism when the budget changes. Figure \ref{fig:s-pricing} reflects the relationship between the budget and the number of NA-canceled order. The MSBA method outperforms the single-stage allocation and  the unified bonus mechanism. Particularly, with a small budget, the MSBA method performs much better than  the single-stage allocation and the unified bonus mechanism . As the budget becomes more and more sufficient, the budget can cover all potential cancel orders, while the rest could not be solved by algorithms. Hence the three algorithms get converged under a larger budget.

%We observe a significant increase in the quantity when the budget increases. Moreover, our single-stage bonus allocation method brings more quantity increase than the unified bonus mechanism. 

\subsubsection{Impact of the number of the allocation stages}

\begin{figure}[tb]
    \centering
    \begin{subfigure}[b]{0.22\textwidth}
    \includegraphics[width=\textwidth]{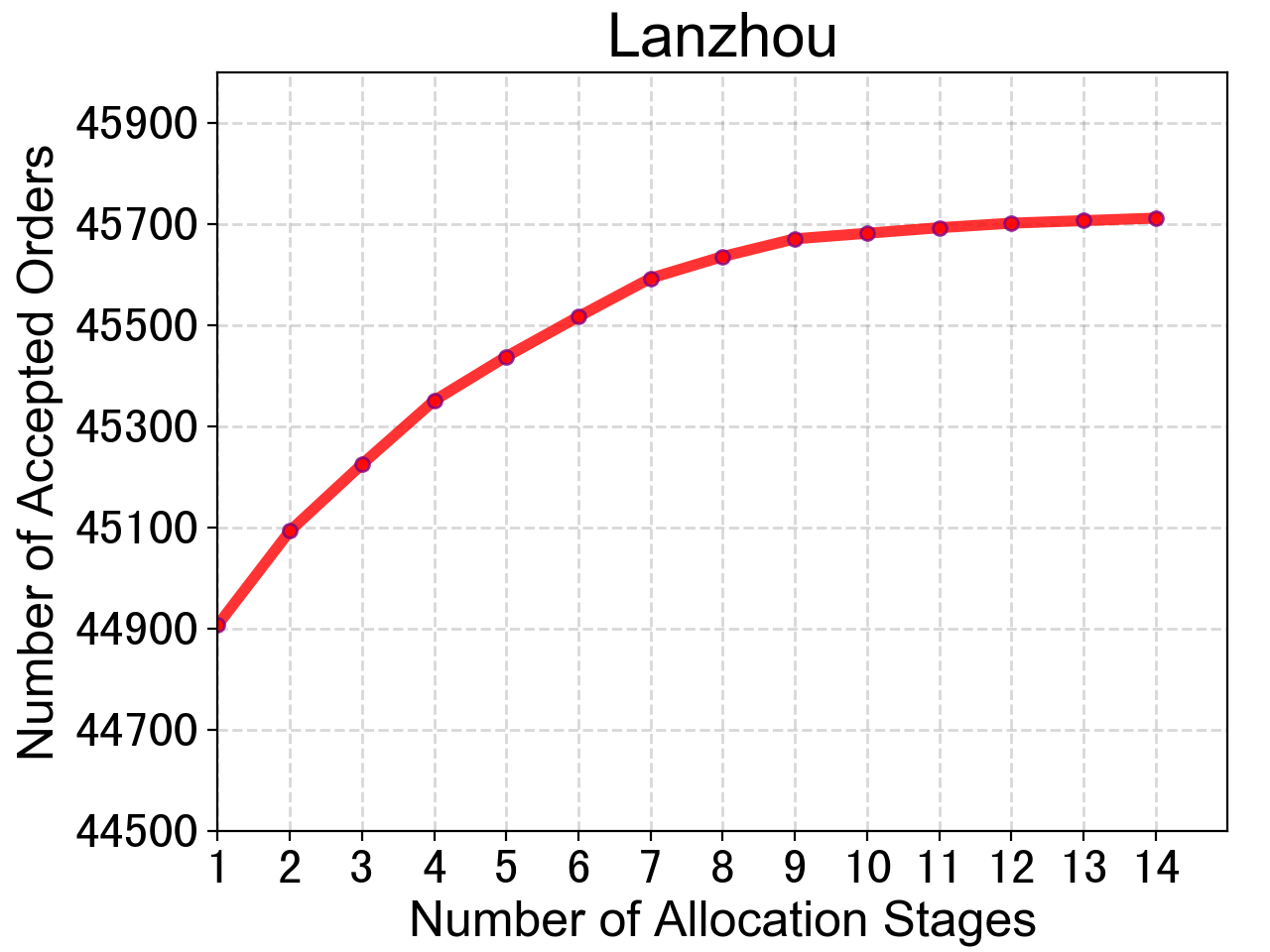}
    \end{subfigure}
    % \quad
    \begin{subfigure}[b]{0.22\textwidth}
    \includegraphics[width=\textwidth]{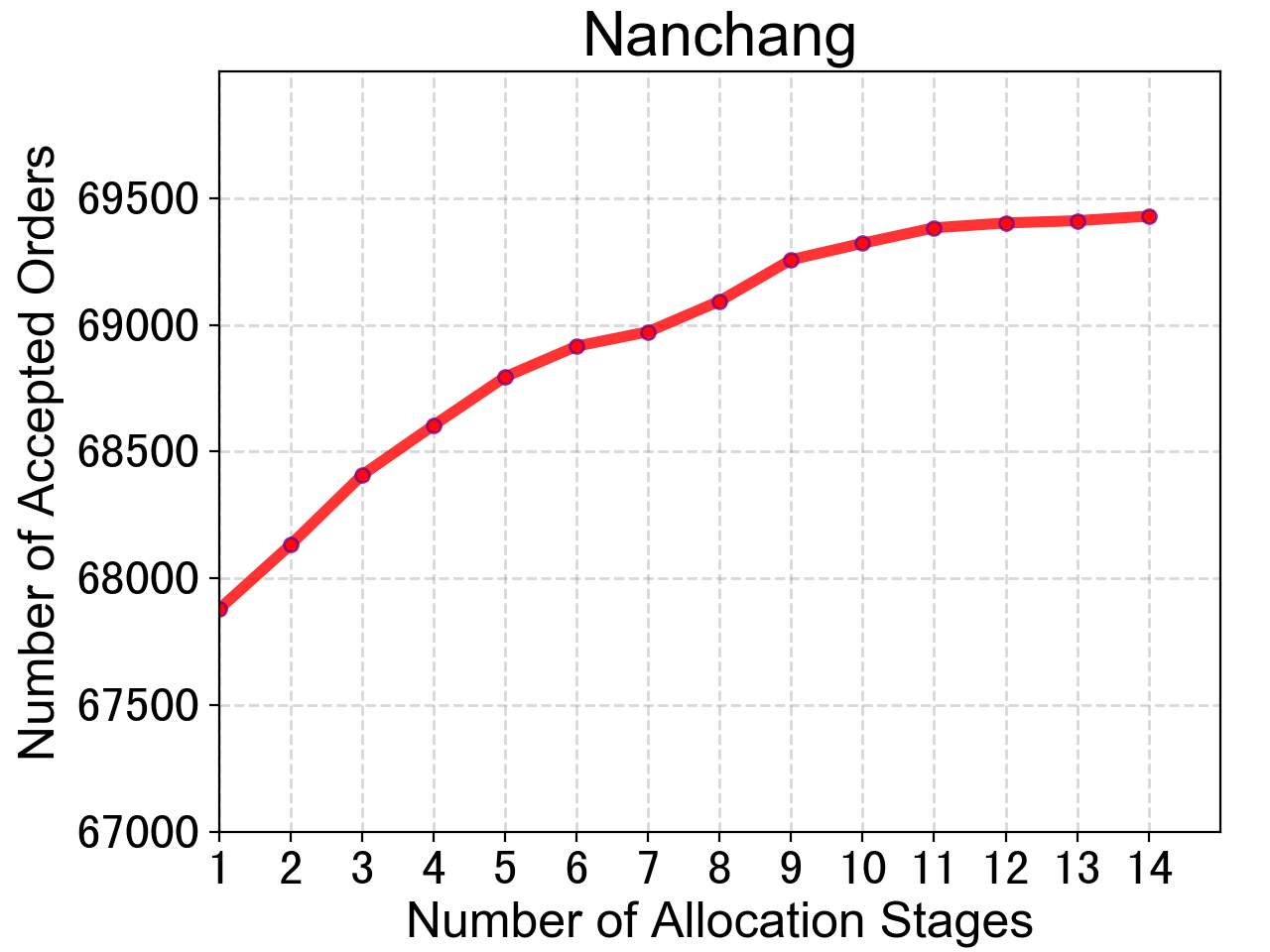}
 
    \end{subfigure}
    
    \begin{subfigure}[b]{0.22\textwidth}
    \includegraphics[width=\textwidth]{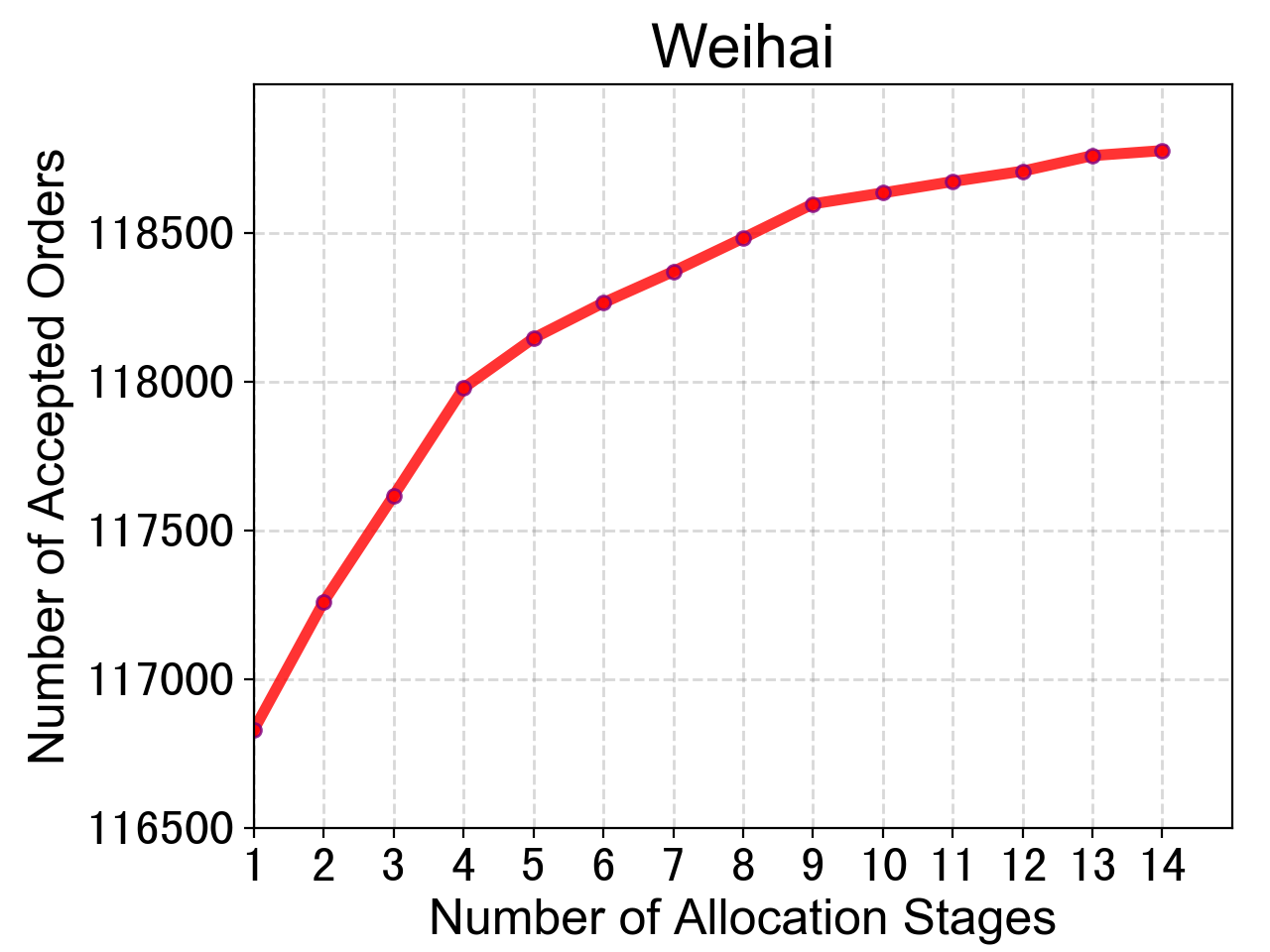}
 
    \end{subfigure}
    \begin{subfigure}[b]{0.22\textwidth}
    % \qquad
    \includegraphics[width=\textwidth]{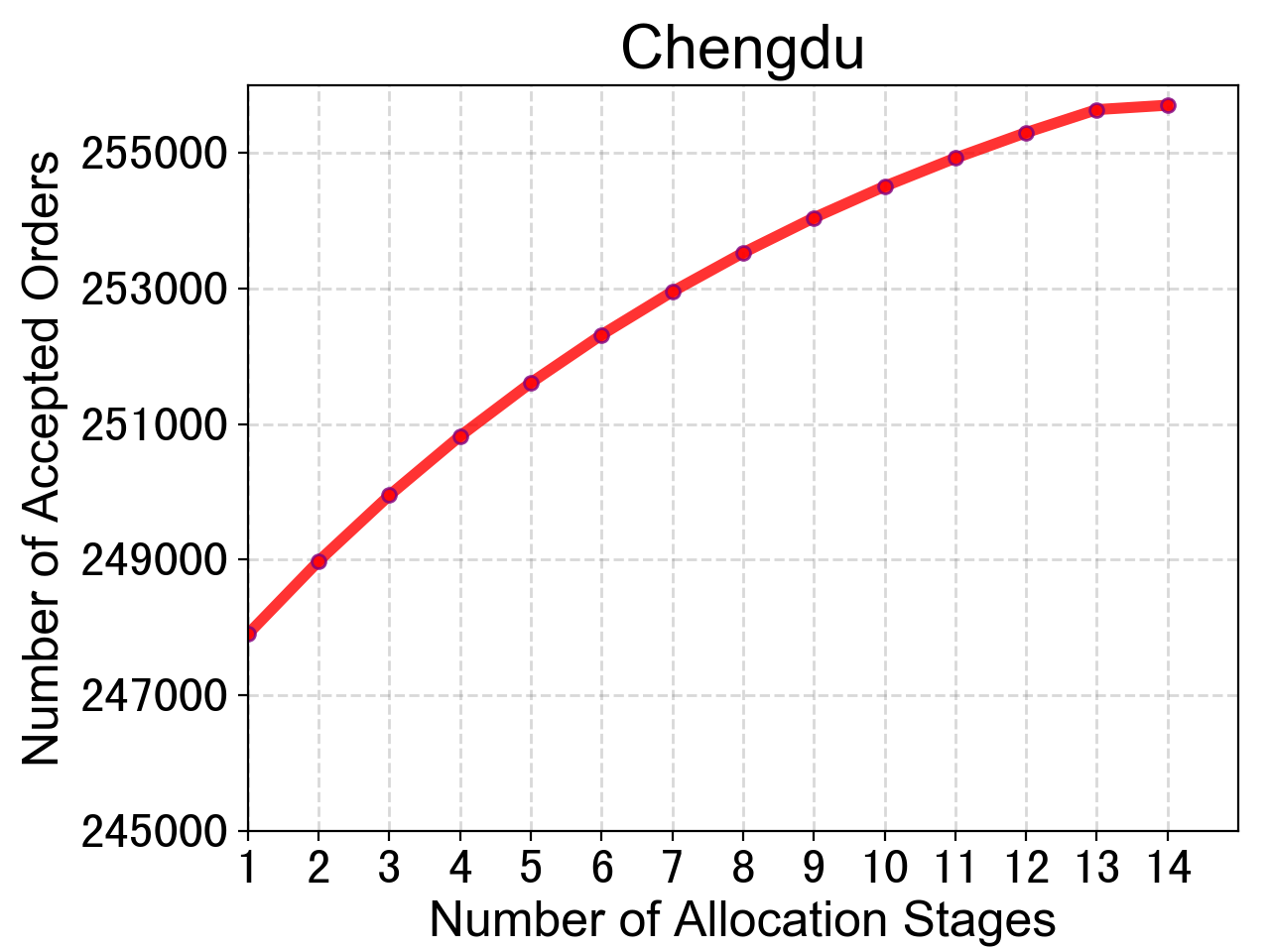}
 
    \end{subfigure}
     \caption{Impact of the number of the allocation stages on the number of accepted orders.}
    \label{fig:stagenumber}
\end{figure}

This section investigates the impact of the number of the allocation stages on the accepted order quantity. As shown in Figure \ref{fig:stagenumber}, the accepted order quantity increases when the number of allocation stages increases. However, the slope of the curve becomes smaller when the number of allocation stages is more than 10. Besides, it is detrimental to the crowdsourcing driver experiences that the delivery bonus of an order changes too frequently. Therefore, to balance the accepted order quantity and the crowdsourcing driver experience, it is reasonable to set the allocation stages no more than 10.

%As an important part of our framework, the acceptance probability model is trained and updated daily using the historical data set over the most recent 30 days.

\subsection{Performance of online A/B tests}

In this section, we introduce our online A/B tests on the Meituan meal delivery platform. 

We implemented our A/B tests on five cities in China, covering 120 delivery areas and 4,960,000 orders per day. The orders are randomly and equally split to 3 groups, adopting multi-stage, single-stage and unified allocation method separately. The budget for both the experimental and control group is set to 0.2 RMB per order. To verify the effectiveness of our algorithm, we recorded the order information data two weeks before and after the start of the experiment for both groups. We employ the LDDP algorithm to calculate the Lagrangian multiplier for each allocation stage by the historical data set 30 days before the experiment. The empirical Lagrangian multipliers are updated every day. The online allocation algorithm is triggered when an order is presented on the screen of the crowdsourcing driver. The delivery bonus of an order is calculated in milliseconds using the empirical Lagrangian multiplier. 

\begin{table*}[t]
\caption{Performance of online A/B tests.}
\label{tab:abtest}
\scalebox{0.90}{
\begin{tabular}{c|cccc|cc|cc}
\hline
\multirow{2}{*}{City Name}  &\multicolumn{4}{c|}{\multirow{1}{*}{Multi-Stage Bonus Allocation}} & \multicolumn{2}{c|}{\multirow{1}{*}{Single-Stage Bonus Allocation}} & \multicolumn{2}{c}{\multirow{1}{*}{Unified Bonus Allocation}} \\
\cline{2-9}
&  \makecell{NA-canceled \\ ratio} & \makecell{drops\% v.s. \\ single-stage} & \makecell{drops\% v.s. \\ unified} & \makecell{ per-order\\ compensation} & \makecell{NA-canceled \\ ratio} & \makecell{per-order\\ compensation}  & \makecell{NA-canceled \\ ratio} & \makecell{per-order\\ compensation}\\
\hline
Nanchang  & 0.68 \%  &	38.47\%  &	60.82\%  &	0.0049 &	1.11\%  &	0.0078	   &	1.74\%  &	0.0086  \\
Shenzhen & 0.49\%&	31.70\%&	 51.62\%&	0.0038 &				0.72\%&	0.0057&			1.01\%&	0.0073\\
Weihai    &0.81\%&		27.51\%	&	40.00\%&		0.0056		&			1.12\%	&	0.0078&					1.35\%&		0.0098 \\
Zhuhai &0.56\%&	 27.06\%&	32.76\%&	0.0038 &				0.77\%&	0.0050&			0.84\%&	0.0061\\
Shantou &1.56\%&	25.02\%&	29.83\%&	0.0073	&				2.09\%&	0.0101&			2.23\%&	0.0113 \\
\hline
\end{tabular}}
\end{table*}

The results are evaluated from two aspects: the NA-canceled order ratio and the compensation paid to restaurants for food waste per order (in RMB), which are two important indicators to the platform. As shown in Table \ref{tab:abtest},  using the proposed method, compared to the single-stage allocation method the NA-canceled orders ratio was reduced by more than 25\%, and more than 29\% compared to the unified allocation method under the same budget. Furthermore, we can save more than 30\% of the compensation paid to restaurants for food waste.

\section{Related Work}\label{sec:Literature}

Some studies have studied dynamic pricing for attended home delivery \cite{klein2018model,koch2020route,yang2016choice} where time slot pricing was implemented to dynamically affect customers' bookings using approximate dynamic programming. \cite{yang2016choice} proposed approximating the opportunity cost by calculating the insertion cost of an incoming request based on the insertion heuristic algorithm. Klein et al.\cite{klein2018model} improved the method of Yang et al. by considering expected future demand, making delivery cost approximation more accurate. Koch et al. \cite{koch2020route} combine and extend the methods in the previous paper by considering more flexible customer choice and dynamic vehicle routing with time windows. Ulmer \cite{ulmer2020dynamic} established a Markov decision process model for dynamic routing and same-day delivery pricing and presented an anticipatory pricing method to solve it. Based on this work of Ulmer \cite{ulmer2020dynamic}, Prokhorchuk et al. \cite{prokhorchuk2019stochastic} considered a stochastic travel time to make the model more applicable. Some references discussed the researches related to the meal delivery problem\cite{yildiz2019provably,ulmer2021restaurant,ding2020delivery}, but few of them involved the pricing and bonus allocation problem. 

There are some studies on dynamic pricing for ride-hailing platforms. Some ride-hailing platforms use a dynamic pricing strategy called surge/prime pricing, in which the base fare is multiplied by a multiplier that is greater than one when the demand is high relative to supply \cite{hall2015effects}. Castillo et al.\cite{castillo2017surge} proposed a steady-state model for dynamic pricing in ride-hailing applications and validated that dynamic pricing is particularly important for ride-hailing owing to the so-called wild "goose chase phenomenon". Zha et al.\cite{zha2018geometric} built a spatial pricing model based on a discrete-time geometric matching framework in which a customer is matched to their closest available vehicle within a set radius. Tong et al. \cite{tong2018dynamic} proposed the Global Dynamic Pricing (GDP) in spatial crowdsourcing in a ride-hailing platform context. It proposes dynamic pricing in multiple local markets with unknown demand, limited supply, or dependent supply. It differs in pricing between ride-hailing and meal delivery problems. For example, dynamic pricing for ride-hailing determines the price presented to passengers spatiotemporally, but a delivery bonus is presented to drivers based on the order features for meal delivery.

\section{Conclusion}\label{sec:conc}

In this paper, we study the MSBA problem for meal delivery platform. We propose a framework including a semi-black-box acceptance probability model, an LDDP algorithm, and an online allocation algorithm to solve this problem. Using this framework, online decision could be made within milliseconds. Both offline experiments and online A/B tests are implemented to verify the effectiveness of our method. The offline experiments show that compared to single-stage bonus allocation and unified bonus mechanism, the number of canceled orders derived from MSBA decrease about 25\% and 50\%, respectively. The online A/B tests show that compared to single-stage bonus allocation the number of canceled orders reduced by more than 25\%. The proposed algorithm has been actually applied in the largest meal delivery platform in China.

\bibliographystyle{ACM-Reference-Format}
\bibliography{sample-base}

\end{document}